\documentclass[11pt]{article}

\usepackage[final]{acl}

\usepackage{times}
\usepackage{latexsym}
\usepackage{amsmath}
\usepackage{amssymb}
\usepackage{subcaption}
\usepackage{enumitem}
\usepackage{tabularx}
\usepackage{booktabs}
\usepackage{multirow}
\usepackage{graphicx}
\usepackage{makecell}
\usepackage{soul}
\usepackage{xcolor}
\usepackage{fontawesome5} 
\usepackage[many]{tcolorbox}
\definecolor{promptPurple}{HTML}{5D1548} 

\newtcolorbox{promptbox}[2][]{%
    enhanced,
    breakable,     
    colback=white,      
    colframe=promptPurple, 
    coltitle=white,     
    fonttitle=\bfseries\large,
    title={\faInfoCircle\ \ #2}, 
    arc=3mm,            
    boxrule=1.5pt,      
    left=5pt, right=5pt, top=5pt, bottom=5pt,
    overlay first={
        \draw[promptPurple, line width=1.5pt] (frame.south west) -- (frame.south east); 
    },
    overlay middle={
        \draw[promptPurple, line width=1.5pt] (frame.north west) -- (frame.north east);
        \draw[promptPurple, line width=1.5pt] (frame.south west) -- (frame.south east);
    },
    overlay last={
        \draw[promptPurple, line width=1.5pt] (frame.north west) -- (frame.north east);
    },
    #1
}

\definecolor{hl_yellow}{HTML}{FFF3A8}
\definecolor{hl_blue}{HTML}{DAE8FC}
\definecolor{hl_pink}{HTML}{FFE3ED}
\definecolor{hl_green}{HTML}{D4F5E3}

\newcommand{\hlb}[1]{%
  {\sethlcolor{hl_blue}\hl{#1}}%
}
\newcommand{\hlp}[1]{%
  {\sethlcolor{hl_pink}\hl{#1}}%
}
\newcommand{\hlg}[1]{%
  {\sethlcolor{hl_green}\hl{#1}}%
}

\usepackage[T1]{fontenc}

\usepackage[utf8]{inputenc}

\usepackage{microtype}

\usepackage{inconsolata}

\usepackage{graphicx}
\usepackage{xspace}

\usepackage{hyperref}
\usepackage{cleveref}
\usepackage{pifont}

\crefname{section}{\S\!}{\S\S\!}
\crefname{table}{Tab.}{Tabs.}
\crefname{figure}{Fig.}{Figs.}
\crefname{algorithm}{Alg.}{Algs.}
\crefname{appendix}{App.}{Apps.}
\crefname{equation}{Eq.}{Eqs.}

\newcommand{\interalia}[1]{\citep[\textit{inter alia}]{#1}}
\newcommand{\method}{\textsc{MentorCollab}\xspace}
\newcommand{\methodfree}{\textsc{MentorCollab-Free}\xspace}
\newcommand{\methodtrain}{\textsc{MentorCollab-MLP}\xspace}

\title{\method: Large-to-Small Inference-Time Mentorship \\for Concise Reasoning in Language Models}

\author{Haojin Wang\thanks{Equal contribution.}\textsuperscript{1} \quad Yike Wang\footnotemark[1]\textsuperscript{2} \quad Shangbin Feng\textsuperscript{2}\\ \textbf{Hannaneh Hajishirzi\textsuperscript{2,3} \quad Yulia Tsvetkov\textsuperscript{2}} \\
                \textsuperscript{1}UIUC \quad \textsuperscript{2}University of Washington\quad  \textsuperscript{3}Allen Institute for Artificial Intelligence
                \\\texttt{haojinw2@illinois.edu} \quad \texttt{yikewang@cs.washington.edu}
                }

\begin{document}
\maketitle
\begin{abstract}
Large reasoning models (LRMs) have demonstrated impressive reasoning capabilities, but their solutions are often verbose and computationally expensive. In contrast, small language models (SLMs) produce concise outputs with lower inference costs, yet they frequently struggle on challenging multi-step reasoning tasks.
Existing inference-time collaboration methods attempt to bridge this gap through \textit{imitation}, encouraging SLMs to follow the reasoning process of LRMs. However, the student often inherits the mentor's overthinking, producing long and reflective reasoning chains while still falling short in accuracy.
We propose \method, a collaboration method based on  \textit{mentorship}: the SLM remains the primary generator and consults the LRM only when additional reasoning support is needed.
At sparsely sampled token positions, we probe for divergence between the two models and use a lightweight verifier to decide whether the SLM should follow a short lookahead segment from its mentor or continue on its own.
Across 15 SLM–LRM pairs and 3 domains, our method achieves an average gain of 3.0\%, with improvements of up to 8.0\% in 12 settings. The resulting traces remain shorter than the mentor's, using only a small fraction of its tokens. These results demonstrate that selective, verified mentorship can boost reasoning accuracy while preserving concise generation.\footnote{Code available at \href{https://github.com/haojinw0027/Mentorcollab}{github.com/haojinw0027/Mentorcollab}.}
\end{abstract}

\section{Introduction}

\begin{figure}[!t]
    \centering
    \begin{subfigure}[t]{\linewidth}
        \centering
        \includegraphics[width=\linewidth]{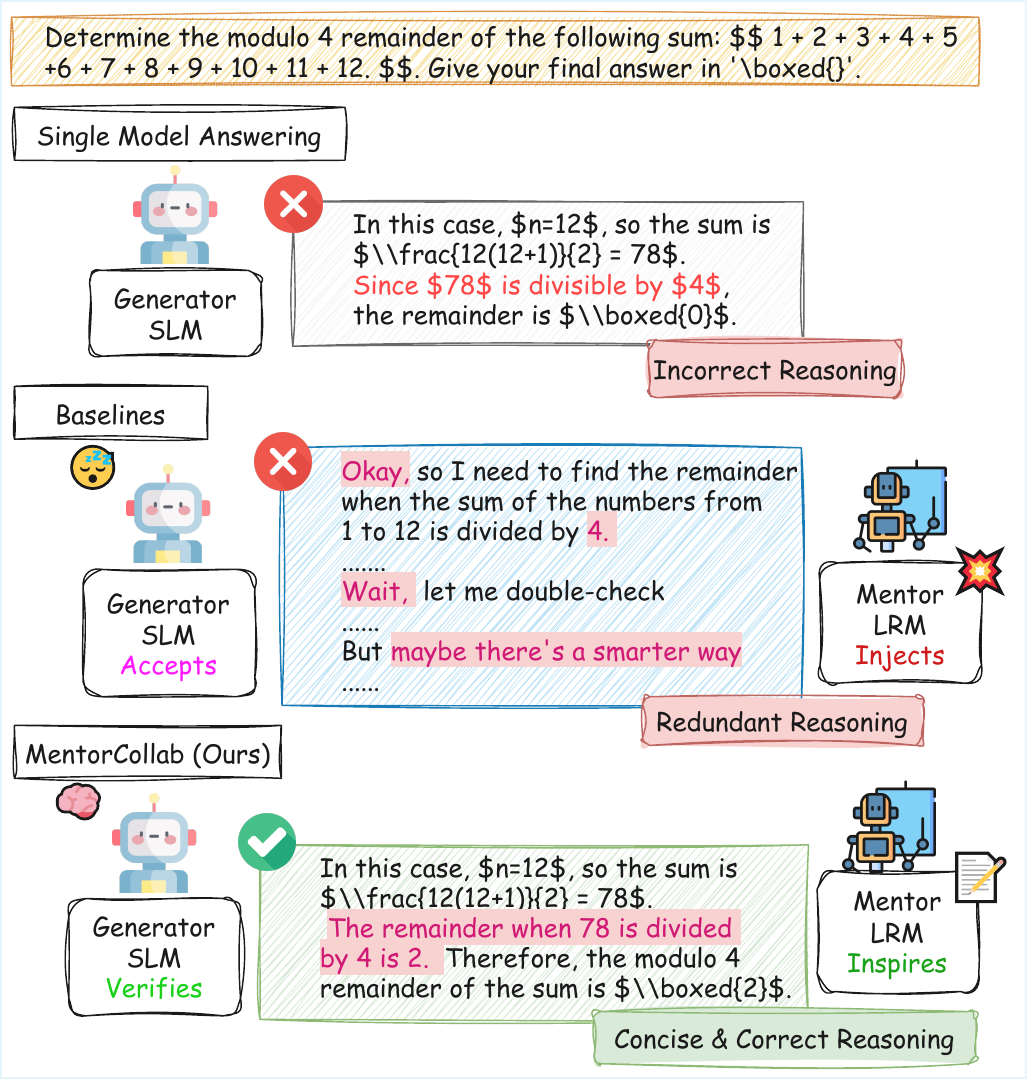}
        \vspace{-10pt}
    \end{subfigure}
    \caption{
    Under \method, the mentor proposes short segments, and the generator verifies rather than heuristically accepts them, achieving correct and concise solutions. }
    \label{fig:teaser_and_budget}
\end{figure}
Large language models trained with long chains of thought achieve strong performance on a wide range of reasoning tasks, from mathematics to commonsense question answering \citep{xu_toward_2025, guo2025deepseek, openai_o3_o4_mini_system_card_2025, moonshotai_kimi_k2_thinking_2025}.
These reasoning-specialized models are effective, but they are also expensive to run and prone to generating redundant or reflective reasoning, leading to high inference cost and overthinking~\interalia{sui2025stop, liu2025diffadapt}. 
In contrast, small language models (SLMs; $\le$8B parameters) are cheaper, faster, and produce concise outputs, but they typically struggle on challenging multi-step reasoning problems \citep{wei2022emergent, fu2023specializing, wang2026small}.
One intuitive solution is to combine their complementary strengths through \emph{model collaboration}~\citep{feng2025one, feng2026moco}. 
In this work, we focus on the inference-time, training-free approaches, which are lightweight and practical as they require no joint optimization or low-level access to model internals.
Existing inference-time approaches either route each query to a single model~\citep{hu2024routerbench, chen2024frugalgpt} or have a stronger model inject tokens into or take over the weaker one's generation.
For example, Nudging~\citep{fei-etal-2025-nudging} injects tokens when the generator is uncertain. 
Adapted to large reasoning models (LRMs), however, such strategies push the small model to \emph{imitate} verbose chain-of-thought, often introducing reflective cues like \emph{``Wait.''}, inflating its traces without reliably correcting errors~\citep{wang2025speculate, chen2025r}. 
Rather than imitation, we argue that the large model should act as a \emph{mentor}: providing brief, targeted cues only when they help, while leaving the small model in control and preserving its conciseness (\cref{fig:teaser_and_budget}). 
We thus seek \emph{selective inference-time mentorship} rather than full takeover.

We introduce \method, an inference-time collaboration that instantiates this idea. 
We refer to the SLM as the \emph{generator} and the LRM as the \emph{mentor}. 
Rather than injecting mentor tokens directly or switching decoding to the mentor, \method randomly samples token positions during generation to probe for divergences between the two models. 
When a divergence is detected, a lightweight verifier decides whether the generator should follow a short lookahead segment from the mentor or continue along its own trajectory. 
This design keeps the generator in charge, while allowing the mentor to provide concise guidance at critical points. 
The verifier comes in two lightweight variants: \methodfree prompts the generator itself, while \methodtrain trains a small MLP once per generator and reuses it across mentors, avoiding the pair-specific joint training required by prior methods~\citep{shen-etal-2024-learning}.

\begin{figure}[!t]
    \centering
    \begin{subfigure}[t]{\linewidth}
        \centering
        \includegraphics[width=\linewidth]{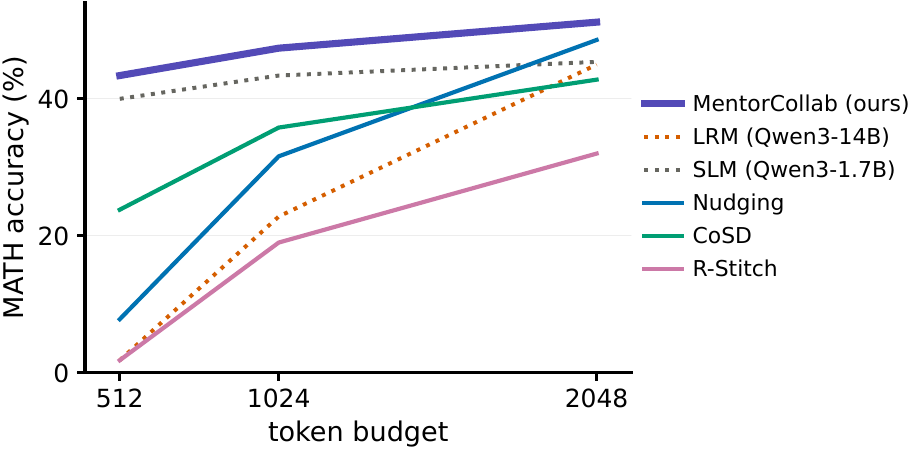}
    \end{subfigure}
    \caption{\method outperforms SLMs, LRMs, and baseline collaboration method under constrained token budgets.}
    \label{fig:teaser_and_budget2}
\end{figure}

We evaluate \method across 15 generator--mentor pairs and three domains: math reasoning, general knowledge, and commonsense reasoning. 
\method improves accuracy in 12 out of 15 settings, achieving an average gain of 3.0\% and peak improvements of up to 8.0\%, while having only 18.4\% of mentor tokens in the final generation on average (\cref{fig:teaser_and_budget2}). 
Further analysis shows that effective collaboration does not require long reasoning traces or frequent mentor intervention: short lookahead segments (4-8 tokens) and moderate probing frequency are sufficient.

Overall, our results show that selective inference-time mentorship can recover much of the reasoning ability of large models at a fraction of the cost. 
This suggests a practical path toward efficient reasoning systems that combine the strengths of large and small language models without requiring additional training or low-level access.

\section{Methodology}

\subsection{Problem Setting}
Consider a SLM $\mathcal{G}$ and a LRM $\mathcal{M}$.
Let $\mathbf{x} = [x_1, x_2, \ldots, x_n] \in \mathbb{Z}_{\geq 0}^n$ denote the input token sequence.
Given $\mathbf{x}$, \method aims to generate a concise yet accurate continuation $[x_{n+1}, \ldots, x_{n+K}]$ through collaborative inference between $\mathcal{G}$ and $\mathcal{M}$.
At each generation step $t$ $(n < t \le n+K)$, we define:
\begin{itemize}
    \item $x_t^{\mathcal{G}}$ be the token generated by $\mathcal{G}$,
    \item $x_t^{\mathcal{M}}$ be the token generated by $\mathcal{M}$ if consulted,
    \item $x_t$ be the final token selected by \method.
\end{itemize}
Given this setup, we next describe how \method dynamically coordinates the generator and mentor during inference time.

\subsection{\method}
\begin{figure*}[!t]
    \includegraphics[width=\linewidth]{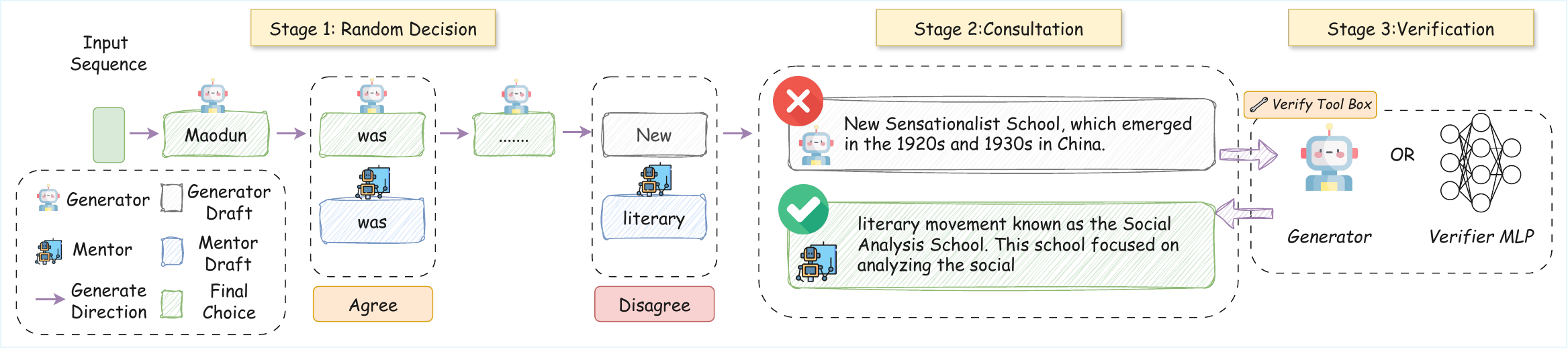}
    \vspace{-15pt}
    \caption{
        Overview of \method. At randomly sampled token positions, if the generator token and the mentor token disagree, we prompt two models to produce a short future segment respectively. A verifier then selects which segment to follow, and generation proceeds from the selected continuation.
    }
    \vspace{-5pt}
    \label{fig:method}
\end{figure*}

\method (\cref{fig:method}) is a lightweight test-time collaboration framework that allows a generator $\mathcal{G}$ to selectively incorporate guidance from a mentor $\mathcal{M}$ during decoding.
The framework operates in three stages—\emph{decision}, \emph{consultation}, and \emph{verification}—which we describe in detail below.

\paragraph{\textbf{Decision}}
To decide whether to consult the mentor, \method independently samples
\begin{align*}
    B_t \sim \mathrm{Bernoulli}(\rho), \rho \in (0,1),
\end{align*}
where $\rho$ is the decision proportion hyperparameter, and queries mentor model $\mathcal{M}$ iff $B_t = 1$.
When a query occurs ($B_t = 1$), both $\mathcal{G}$ and $\mathcal{M}$ propose their next tokens $x_t^{\mathcal{G}}$ and $x_t^{\mathcal{M}}$.
Following implementations from \citet{fei-etal-2025-nudging}, such comparison is performed at whitespace boundaries, effectively comparing at the word level.
We define a disagreement indicator $D_t$ as follows:
\begin{align*}
    D_t = \begin{cases}
        1 & \text{if } x_t^{\mathcal{G}} \neq x_t^{\mathcal{M}}, \\
        0 & \text{otherwise}.
    \end{cases}
\end{align*}
and \method proceeds to the next stage of consultation only when $D_t = 1$.

\paragraph{\textbf{Consultation}}
Once disagreement is detected ($D_t = 1$), \method prompts both $\mathcal{G}$ and $\mathcal{M}$ to produce short future segment $\mathbf{s}$.
For each model $m \in \{\mathcal{G}, \mathcal{M}\}$, we define its proposed segment as:
\begin{align*}
    \mathbf{s}^m = [x_t^m, x_{t+1}^m, \ldots, x_{t+L-1}^m],
\end{align*}
where $L$ is the length of the segment.

\paragraph{\textbf{Verification}}
Given the proposed segments $\mathbf{s}^{\mathcal{G}}$ and $\mathbf{s}^{\mathcal{M}}$, \method employs a lightweight verifier $\mathcal{V}$ to select between them.
We consider two verifier variants: \methodfree and \methodtrain.

\textbf{\methodfree} We directly use the generator $\mathcal{G}$ itself as the verifier $\mathcal{V}_{\text{free}}$ through a verification prompt.
Given the two candidate segments, we append a short query asking which segment should be chosen (e.g., \textit{“Now I will choose the next sequence that could lead to the correct answer. (Option A or B?)”}), without revealing which model produced which segment.
The verifier $\mathcal{V}_{\text{free}}$ outputs a choice (“A” or “B”), which we map to a decision $\hat{m}_t \in \{\mathcal{G}, \mathcal{M}\}$ indicating which segment to follow.

\textbf{\methodtrain} Instead of directly treating $\mathcal{G}$ as the verifier, we train a lightweight multi-layer perceptron (MLP) verifier $\mathcal{V}_{\text{train}}$.
To learn a stable verifier, we formalize the training of $\mathcal{V}_{\text{train}}$ as a supervised binary classification problem.
Running \methodfree over a training split $\mathcal{D}_{\text{train}}$ with model pair $(\mathcal{G}, \mathcal{M})$, we obtain a dataset $\mathcal{T}=\{(\mathbf{h}_i^{\mathcal{G}}, \hat{m}_i)\}_{i=1}^N$, where $\mathbf{h}_i^{\mathcal{G}}$ is the last-layer hidden state of $\mathcal{G}$ at the verification index $i$ and $\hat{m}_i \in \{\mathcal{G}, \mathcal{M}\}$ is the decision which leads to the correct continuation.
We curate $\mathcal{T}$ for samples where the generator $\mathcal{G}$ fails but \methodfree succeeds by choosing the correct segment.
The verifier $\mathcal{V}_{\text{train}}$ is then formalized as $f_\theta: \mathbb{R}^d \to \{0, 1\}$, where $d$ is the dimension of the hidden state.
The verifier $f_\theta$ outputs $f_\theta(\mathbf{h}_i^{\mathcal{G}})$ interpreted as $\mathbb{P}$.
We train $\theta$ by minimizing the empirical binary cross-entropy loss:
\begin{align*}
    \begin{aligned}
    \mathcal{L}(\theta)
    &= -\frac{1}{N} \sum_{i=1}^N 
        \Bigl[ \hat{m}_i \log f_\theta(\mathbf{h}_i^{\mathcal{G}})
    \\[0.25em]
    &\qquad
        + (1-\hat{m}_i)\log\bigl(1 - f_\theta(\mathbf{h}_i^{\mathcal{G}})\bigr) \Bigr].
    \end{aligned}
    \end{align*}
using mini-batch Adam \cite{DBLP:journals/corr/KingmaB14} and validation-based early stopping.
At test time, the learned verifier $f_\theta$ replaces the heuristic decision rule by selecting the mentor branch whenever $f_\theta(\mathbf{h}_i^{\mathcal{G}}) > 0.5$.
We give a detailed description of the verifier training in \cref{app:verifier_training}.

\section{Experiments}

\subsection{Experiment Setup}
\paragraph{\textbf{Generators and Mentors}}
We select three small base models (Llama3.1-8B~\cite{DBLP:journals/corr/abs-2407-21783}, Gemma3-4B-PT~\cite{DBLP:journals/corr/abs-2503-19786}, and Qwen3-8B-Base~\cite{DBLP:journals/corr/abs-2505-09388}) and two small instruct models (Llama3.2-3B-Instruct and Qwen3-1.7B) from diverse model families and sizes as the generators.
we select three representative LRMs: Qwen3-14B and Qwen3-32B from the Qwen3 family \cite{DBLP:journals/corr/abs-2505-09388}, and DeepSeek-R1-Distilled-Llama-70B from the R1 distilled family \cite{guo2025deepseek}.

\paragraph{Hyperparameters}
To balance inference cost and generation quality, we conduct our main experiments with a decision proportion of $\rho = 25\%$ and a segment size of $L = 16$.
As SLMs typically produce concise reasoning traces, we cap the maximum generation length at 512 tokens across all experiments and adopt greedy decoding.

\begin{table*}[!t]
  \centering
  \small

  \resizebox{\textwidth}{!}{%
  \begin{tabular}{llcc|cccccc}
      \toprule
      \textbf{Generator Name} &
      \textbf{Mentor Name} &
      \textbf{Generator} &
      \textbf{Mentor} &
      \textbf{\makecell{Average\\ Decoding}} &
      \textbf{\makecell{Nudging\\($\gamma=0.40$)}} &
      \textbf{\makecell{CoSD\\($\alpha=\beta=0.50$)}} &
      \textbf{\makecell{R-Stitch\\($\tau=0.03$)}} &
      \textbf{\makecell{\method \\ \textsc{FREE} \\($\rho=25\%$)}} &

      \textbf{\makecell{\method \\ \textsc{MLP} \\($\rho=25\%$)}} \\
      \midrule        
      \multicolumn{10}{c}{\textit{Domain1: Math Reasoning} (\textbf{MATH})} \\
      \midrule
  \multirow{3}{*}{Gemma-3-4B-PT}
  & Qwen3-14B      & \multirow{3}{*}{14.20\%} & 1.80\%
  & 3.40\% & 8.00\% &  5.40\% &  3.00\% & \underline{15.80\%} & \textbf{21.00\%}  \\
  & Qwen3-32B      &                         & 3.20\%
  & 4.60\% & 9.80\% &  8.00\% &  1.80\% & \underline{14.60\%} & \textbf{20.80\%} \\
  & R1-Distilled-Llama-70B &                  & 21.00\%
  & \underline{21.60\%}  & 5.20\% & 11.40\% & 17.00\% & 15.00\% & \textbf{22.00\%} \\
  \midrule

  \multirow{3}{*}{Llama3.1-8B}
  & Qwen3-14B      & \multirow{3}{*}{12.20\%} &  1.80\%
  & 2.80\% & \underline{17.80\%} & 12.80\% & 1.80\%   & 13.60\% & \textbf{18.20\%} \\
  & Qwen3-32B      &                         &  3.20\%
  & 2.80\% & \underline{17.20\%} & 15.00\% & 2.60\%  & 13.20\% & \textbf{18.00\%} \\
  & R1-Distilled-Llama-70B &                  & 21.00\%
  & \textbf{21.00\%} & 13.20\% & 14.80\% & \underline{19.20\%}  & 12.00\% & 15.80\% \\
  \midrule

  \multirow{3}{*}{Qwen3-8B-Base}
  & Qwen3-14B      & \multirow{3}{*}{43.60\%} &  1.80\%
  & 4.00\% & 26.40\% & 23.80\% &  2.20\% & \underline{44.40\%} & \textbf{46.60\%} \\
  & Qwen3-32B      &                         &  3.20\%
  & 6.40\% & 26.00\% & 27.00\% &  3.00\% & \underline{45.80\%} & \textbf{46.80\%} \\
  & R1-Distilled-Llama-70B &                  & 21.00\%
  & 23.60\% & 14.40\% & 27.80\% & 16.40\% & \underline{40.20\%} & \textbf{42.40\%} \\
  \midrule

  \multirow{3}{*}{Llama3.2-3B-Instruct}
  & Qwen3-14B      & \multirow{3}{*}{26.00\%} & 1.80\%
  & 4.00\% & 28.20\% &  22.80\% &  2.00\% & \underline{29.60\%} & \textbf{31.80\%}  \\
  & Qwen3-32B      &                         & 3.20\%
  & 6.40\% & \underline{30.60\%} &  25.80\% &  2.60\% & 30.00\% & \textbf{31.00\%} \\
  & R1-Distilled-Llama-70B &                  & 21.00\%
  & \underline{31.60\%} & 28.20\% & 30.00\% & 19.20\% & 27.80\% & \textbf{33.20\%} \\
  \midrule

  \multirow{3}{*}{Qwen3-1.7B}
  & Qwen3-14B      & \multirow{3}{*}{40.00\%} & 1.80\%
  & 11.20\% & 7.80\% &  23.80\% &  1.80\% & \underline{43.20\%} & \textbf{43.40\%}  \\
  & Qwen3-32B      &                         & 3.20\%
  & 9.80\% & 8.80\% &  23.40\% &  2.80\% & \underline{44.00\%} & \textbf{45.00\%} \\
  & R1-Distilled-Llama-70B &                  & 21.00\%
  & 31.80\% & 13.60\% & 26.40\% & 18.00\% & \textbf{44.00\%} & \underline{43.60\%} \\
  \midrule
  \multicolumn{10}{c}{\textit{Domain2: General Knowledge} (\textbf{SuperGPQA})} \\
  \midrule

  \multirow{3}{*}{Gemma-3-4B-PT}
    & Qwen3-14B      & \multirow{3}{*}{13.80\%} & 11.80\%
    & 9.60\% & 8.60\% & 6.40\% & 11.20\% & \underline{12.60\%} & \textbf{15.20\%} \\
    & Qwen3-32B   &                         & 10.60\%
    & 3.60\% & 8.80\% & 2.60\% & 9.80\% & \textbf{16.00\%} & \underline{14.40\%} \\
    & R1-Distilled-Llama-70B      &                         & 17.40\%
    & \textbf{18.20\%} & 11.80\% & 11.80\% & \underline{16.60\%} & 15.20\% & 14.80\% \\
  \midrule

  \multirow{3}{*}{Llama3.1-8B}
    & Qwen3-14B      & \multirow{3}{*}{18.00\%} & 11.80\%
    & 8.80\% & 11.40\% & 7.80\% & 10.80\% & \textbf{17.00\%} & \textbf{17.00\%} \\
    & Qwen3-32B   &                         & 10.60\%
    & 2.80\% & 8.80\% & 2.80\% & 10.20\% & \underline{16.60\%} & \textbf{17.40\%} \\
    & R1-Distilled-Llama-70B      &                         & 17.40\%
    & \textbf{21.0\%} & 11.40\% & 16.40\% & 16.60\% & \underline{16.80\%} & 15.40\% \\
  \midrule

  \multirow{3}{*}{Qwen3-8B-Base}
    & Qwen3-14B      & \multirow{3}{*}{15.40\%} & 11.80\%
    & 13.20\% & 11.40\% & 15.40\% & 10.20\% & \textbf{17.20\%} & \underline{16.80\%} \\
    & Qwen3-32B   &                         & 10.60\%
    & 13.00\% & 10.00\% & 15.20\% & 10.40\% & \underline{16.20\%} & \textbf{16.80\%} \\
    & R1-Distilled-Llama-70B      &                         & 17.40\%
    & \textbf{18.60\%} & 14.00\% & 12.80\% & 15.80\% & \textbf{18.60\%} & 17.60\% \\
  \midrule

  \multirow{3}{*}{Llama3.2-3B-Instruct}
  & Qwen3-14B      & \multirow{3}{*}{13.20\%} & 11.80\%
  & \textbf{14.80\%} & 12.00\% &  12.00\% &  9.80\% & \underline{13.40\%} & 13.20\%  \\
  & Qwen3-32B      &                         & 10.60\%
  & 9.20\% & 9.40\% &  10.80\% &  10.20\% & \underline{15.00\%} & \textbf{15.80\%} \\
  & R1-Distilled-Llama-70B &                  & 17.40\%
  & \textbf{20.40\%} & 16.00\% & 14.60\% & 16.20\% & 14.20\% & \underline{17.00\%} \\
  \midrule

  \multirow{3}{*}{Qwen3-1.7B}
  & Qwen3-14B      & \multirow{3}{*}{13.60\%} & 11.80\%
  & 14.00\% & 11.00\% &  12.80\% &  11.40\% & \textbf{15.40\%} & \underline{14.20\%}  \\
  & Qwen3-32B      &                         & 10.60\%
  & 10.60\% & 8.40\% &  12.20\% &  9.60\% & \underline{12.40\%} & \textbf{15.80\%} \\
  & R1-Distilled-Llama-70B &                  & 17.40\%
  & \underline{18.00\%} & 15.00\% & 12.80\% & 15.20\% & 12.60\% & \textbf{18.60\%} \\
  \midrule
  \multicolumn{10}{c}{\textit{Domain3: Commonsense Reasoning} (\textbf{Com$^2$-hard Intervention})} \\
  \midrule

  \multirow{3}{*}{Gemma-3-4B-PT}
    & Qwen3-14B      & \multirow{3}{*}{24.07\%} & 2.49\%
    & 3.73\% & 2.90\% & 0.83\% & 4.98\% & \underline{23.24\%} & \textbf{25.73\%} \\
    & Qwen3-32B   &                         & 8.30\%
    & 7.47\% & 9.96\% & 3.73\% & 7.88\% & \underline{21.99\%} & \textbf{23.24\%} \\
    & R1-Distilled-Llama-70B      &                         & 2.49\%
    & 5.39\% & 7.47\% & 2.49\% & 1.66\% & \underline{19.09\%} & \textbf{26.97\%} \\
  \midrule

  \multirow{3}{*}{Llama3.1-8B}
    & Qwen3-14B      & \multirow{3}{*}{30.29\%} & 2.49\%
    & 3.73\% & 2.49\% & 0.83\% & 2.49\% & \underline{31.54\%} & \textbf{33.61\%} \\
    & Qwen3-32B   &                         & 8.30\%
    & 9.96\% & 11.62\% & 3.73\% & 7.88\% & \textbf{30.71\%} & \underline{30.29\%} \\
    & R1-Distilled-Llama-70B      &                         & 2.49\%
    & 6.64\% & 4.98\% & 2.90\% & 2.07\% & \textbf{32.78\%} & \underline{26.56\%} \\
  \midrule

  \multirow{3}{*}{Qwen3-8B-Base}
    & Qwen3-14B      & \multirow{3}{*}{54.77\%} & 2.49\%
    & 3.73\% & 18.67\% & 7.05\% & 2.90\% & \textbf{51.87\%} & \underline{42.32\%}\\
    & Qwen3-32B   &                         & 8.30\%
    & 12.45\% & 21.33\% & 15.77\% & 9.54\% & \textbf{49.79\%} & \textbf{49.79\%} \\
    & R1-Distilled-Llama-70B      &                         & 2.49\%
    & 35.27\% & 22.41\% & 15.77\% & 3.32\% & \textbf{54.77\%} & \underline{48.55\%} \\
  \midrule
  \multirow{3}{*}{Llama3.2-3B-Instruct}
  & Qwen3-14B      & \multirow{3}{*}{53.11\%} & 2.49\%
  & 4.56\% & 7.88\% &    2.90\% & 2.07\% & \textbf{50.21\%} & \underline{40.25\%}  \\
  & Qwen3-32B      &                         & 8.30\%
  & 12.03\% & 19.50\% &    8.30\% & 7.88\% & \textbf{47.72\%} & \underline{42.74\%} \\
  & R1-Distilled-Llama-70B &                  & 2.49\%
  & 14.11\% & 18.67\% &  7.88\% & 2.07\% & \textbf{48.96\%} & \underline{38.17\%} \\
  \midrule
  \multirow{3}{*}{Qwen3-1.7B}
  & Qwen3-14B      & \multirow{3}{*}{9.13\%} & 2.49\%
  & 0.83\% & 1.24\% &    2.49\% & 2.49\% & \underline{9.54\%} & \textbf{9.96\%}  \\
  & Qwen3-32B      &                         & 8.30\%
  & 2.90\% & 0.41\% &    0.83\% & 6.64\% & \underline{9.96\%} & \textbf{13.69\%} \\
  & R1-Distilled-Llama-70B &                  & 2.49\%
  & 1.66\% & 2.49\% &  1.24\% & 5.39\% & \underline{7.47\%} & \textbf{8.30\%} \\
  \midrule
  \end{tabular}%
  } 
  
  \caption{\method on 15 generator–mentor pairs across three domains. We cap all methods at 512 tokens and use greedy decoding throughout. Even though the mentor model collapses under such a tight token budget, for 12 out of 15 cases as shown in \cref{app:average}, \method still leverages it as effective guidance, helping the generator correct its initial mistakes. We bold the best performance and underline the second best performance for each generator–mentor pair.}
  \label{tab:collab_results}
  \end{table*}

\paragraph{\textbf{Domains and Datasets}}
We evaluate \method across three domains: MATH \citep{hendrycksmath2021} for mathematical reasoning, SuperGPQA \citep{DBLP:journals/corr/abs-2502-14739} for general knowledge, and Com$^2$-hard-Intervention \citep{xiong-etal-2025-com2} for commonsense reasoning.
MATH contains competition-level mathematics problems, and we use its official test split.
SuperGPQA covers graduate-level knowledge across 285 different disciplines; we randomly sample 500 test examples.
Com$^2$-hard-Intervention targets challenging commonsense reasoning via causal interventions, and we evaluate on 241 single-answer instances.
For \methodtrain, the verifier is trained on the MATH training split for mathematical reasoning.
For general knowledge and commonsense reasoning, it is trained on a shared set of 4,000 examples sampled from SuperGPQA.
We use 4-shot prompts for MATH, 5-shot for SuperGPQA, and zero-shot for Com$^2$-hard-Intervention, with all prompt templates provided in \cref{app:prompts}.

\paragraph{\textbf{Baselines}}
We compare \method with five baselines, either with or without training.
\begin{itemize}
\item \textbf{Average Decoding}, which average the top-5 token distributions of two models at each token position and select the token with the highest probability in the averaged distribution, following \citet{fei-etal-2025-nudging}.
\item \textbf{Nudging}~\citep{fei-etal-2025-nudging}, which hands over generation to the mentor whenever the generator’s next-token probability falls below a confidence threshold.
\item \textbf{CoSD}~\citep{wang2025speculate}, which employs a rule-based verification module in which the generator hands over to the mentor when the generator is uncertain and the mentor is sufficiently confident.
\item \textbf{R-Stitch}~\citep{chen2025r}, which uses distribution entropy as a proxy signal to decide when to switch between models, switching to the mentor when the generator’s entropy is high and back when it is low.
\item \textbf{Co-LLM}~\citep{shen-etal-2024-learning}, which trains a single layer to classify the hidden state of the generator. We evaluate Co-LLM only on three generator–mentor pairs with compatible tokenizers and only on the MATH domain, since the method depends on tokenizer compatibility and supervised training on required ground-truth training sequences, which significantly increases evaluation cost and limits scalability across diverse model pairs.
\end{itemize}

\subsection{Results}
We report the results of \method on multiple generator–mentor model pairs across different domains in \cref{tab:collab_results}.
Under the short token budget in our setting, mentor LRMs frequently collapse mid-reasoning, consistent with prior empirical observations \citep{srivastava2025llms}.
Prior collaboration methods align generators with mentors at the token level, often inheriting verbose or over-extended reasoning that sacrifices the conciseness of the SLM.
In contrast, \method performs segment-level selection, incorporating mentor guidance only when it corrects the generator’s initial errors while preserving the generator’s concise reasoning style.
Additional analysis over averaged results are presented in \cref{app:average}.
Across three domains and mentor configurations, our selected small generators outperform their single-model baselines in 12 out of 15 cases, typically improving by 3–8 absolute points. 
In contrast, prior collaboration baselines rarely yield more than 1–2 points and often underperform the generator itself.
We further compare \methodtrain with the training-based Co-LLM~\citep{shen-etal-2024-learning}; full results are presented in \cref{app:training-based}.

Beyond the 512-token budget, to verify that these gains are not an artifact of the tight cap, we extend the comparison to 1024- and 2048-token budgets on MATH (\cref{tab:larger_budget}). \method remains the best or competitive across most pairs, and although the mentor recovers accuracy as the budget grows, it does so through longer, backtracking-prone traces, whereas \method keeps the generator's concise trajectory. The one exception is Qwen3-8B+14B at 2048, where the extended budget lets the mentor's verbose reasoning complete and Nudging benefits accordingly; here \method stays on par with the generator baseline.

\begin{table*}[!t]
  \centering
  \small
  \begin{tabular}{llccccc}
  \toprule
  \textbf{Budget} & \textbf{Model Pair} & \textbf{Generator} & \textbf{Mentor} & \textbf{Nudging} & \textbf{CoSD} & \textbf{\makecell{\textsc{Mentor-}\\\textsc{Collab}}} \\
  \midrule
  \multirow{3}{*}{1024}
  & Qwen3-1.7B + 14B & 43.40\% & 22.80\% & 31.60\% & 35.20\% & \textbf{47.40\%} \\
  & Qwen3-1.7B + 32B & 43.40\% & 26.80\% & 33.80\% & 33.00\% & \textbf{51.80\%} \\
  & Qwen3-8B + 14B & 44.80\% & 22.80\% & 44.00\% & 42.40\% & \textbf{48.40\%} \\
  \midrule
  \multirow{3}{*}{2048}
  & Qwen3-1.7B + 14B & 45.40\% & 45.00\% & 48.60\% & 43.80\% & \textbf{51.20\%} \\
  & Qwen3-1.7B + 32B & 45.40\% & 49.40\% & 48.20\% & 42.60\% & \textbf{51.40\%} \\
  & Qwen3-8B + 14B & 44.20\% & 45.00\% & \textbf{56.20\%} & 49.40\% & 45.40\% \\
  \bottomrule
  \end{tabular}
  \caption{Accuracy on MATH under larger token budgets (1024 and 2048).}
  \label{tab:larger_budget}
\end{table*}

\begin{figure}[!t]
  \centering
  \includegraphics[width=0.9\linewidth]{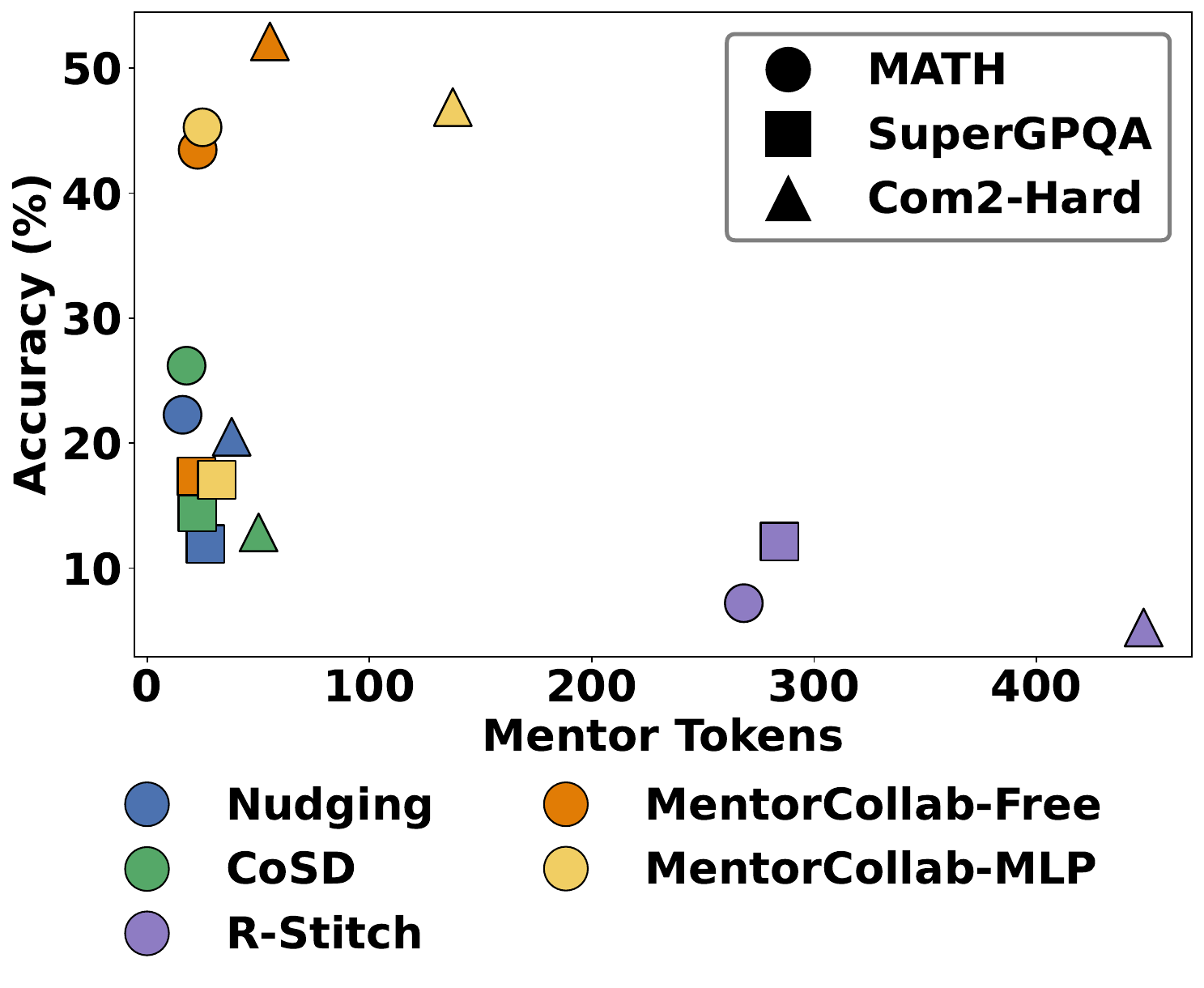}
  \caption{\textbf{Mentor token usage vs.\ accuracy on Qwen3-8B-Base.} \method reaches higher accuracy than baselines while incorporating comparable or fewer mentor tokens.}
  \label{fig:token-perf}
\end{figure}

Furthermore, \method achieves these gains with incorporated token usage comparable to other baselines, as shown in \cref{fig:token-perf}.
By injecting only 10–30\% of the mentor’s tokens, \method improves the small model’s performance by selectively injecting short mentor reasoning segments that inspire the generator’s own inference, rather than overwhelming it with full mentor trajectories.
Overall, the findings suggest that \method provides a stable and efficient way to collaborate small-scale models with LRMs, offering a new perspective on the complementarity of these two model types.

\section{Analysis}

\subsection{Decision Proportion $\rho$}
\begin{figure}[t]
    \centering
    \includegraphics[width=\linewidth]{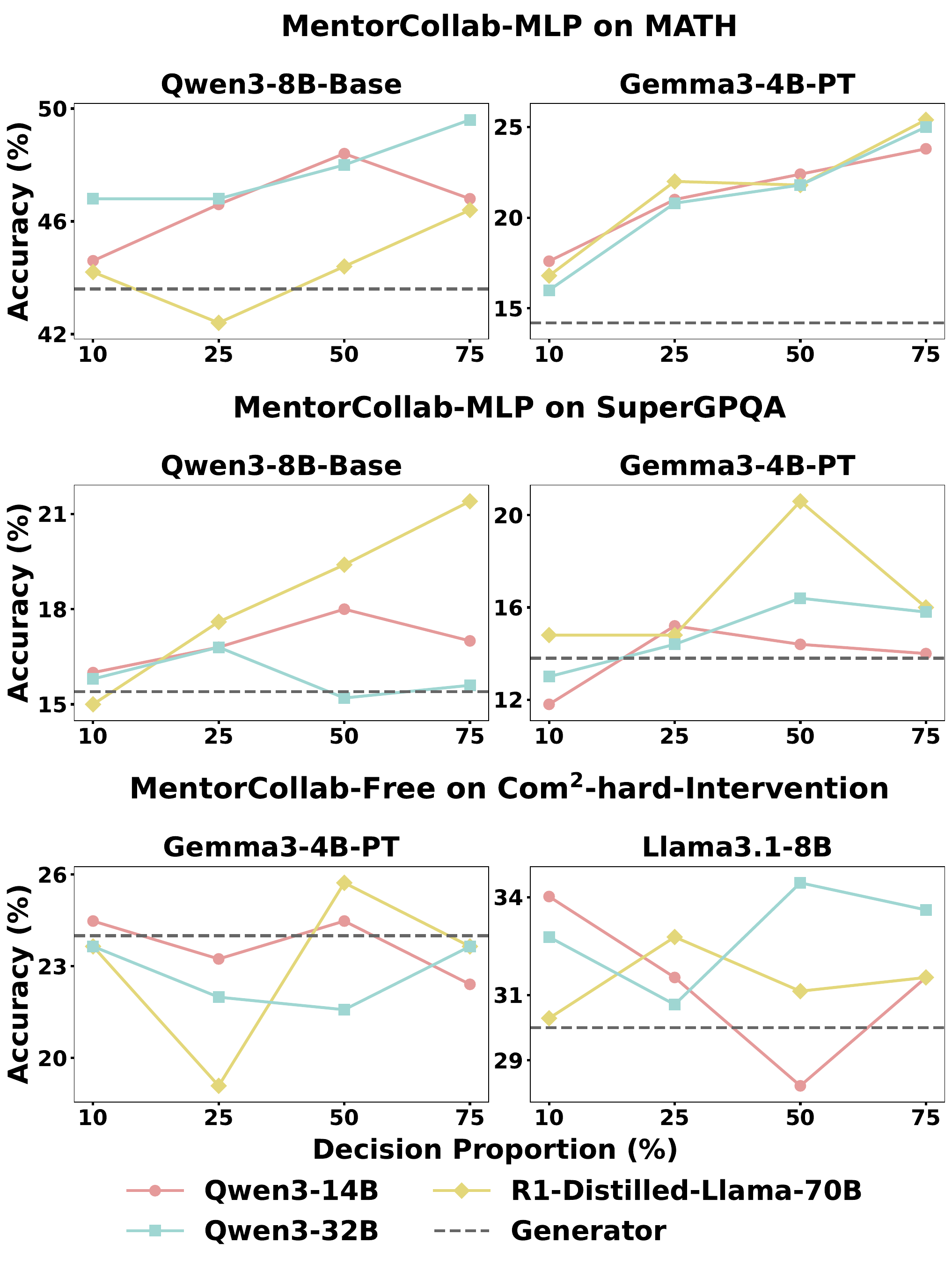}
    \caption{\method performance with different decision proportions.
    We report the performance from multiple generator models with multiple mentor LRMs.
    }
    \label{fig:decision_proportion}
\end{figure}

In this section, we further examine how varying the decision proportion $\rho$ affects the effectiveness of \method.
\Cref{fig:decision_proportion} presents results on MATH and SuperGPQA under different values of~$\rho$.
On MATH, the performance of \methodtrain consistently improves as $\rho$ increases, suggesting that more frequent incorporation of the mentor’s reasoning benefits the generator model.
In contrast, on SuperGPQA, larger $\rho$ does not necessarily yield higher accuracy, and performance can decline when $\rho$ becomes too large.
As in \cref{tab:collab_results}, we empirically find \methodfree performs better than \methodtrain on Com$^2$-hard-Intervention, we also conduct the decision proportion analysis on \methodfree on this domain.
We find that decision proportion does not improve the performance of \methodfree, and introducing more decisions could lead to performance degradation.
This indicates that the decision proportion is domain- and model-dependent, and increasing it leads to more frequent mentor intervention, but such additional guidance does not necessarily leads to better results.

\begin{figure}[t]
    \centering
    \includegraphics[width=\linewidth]{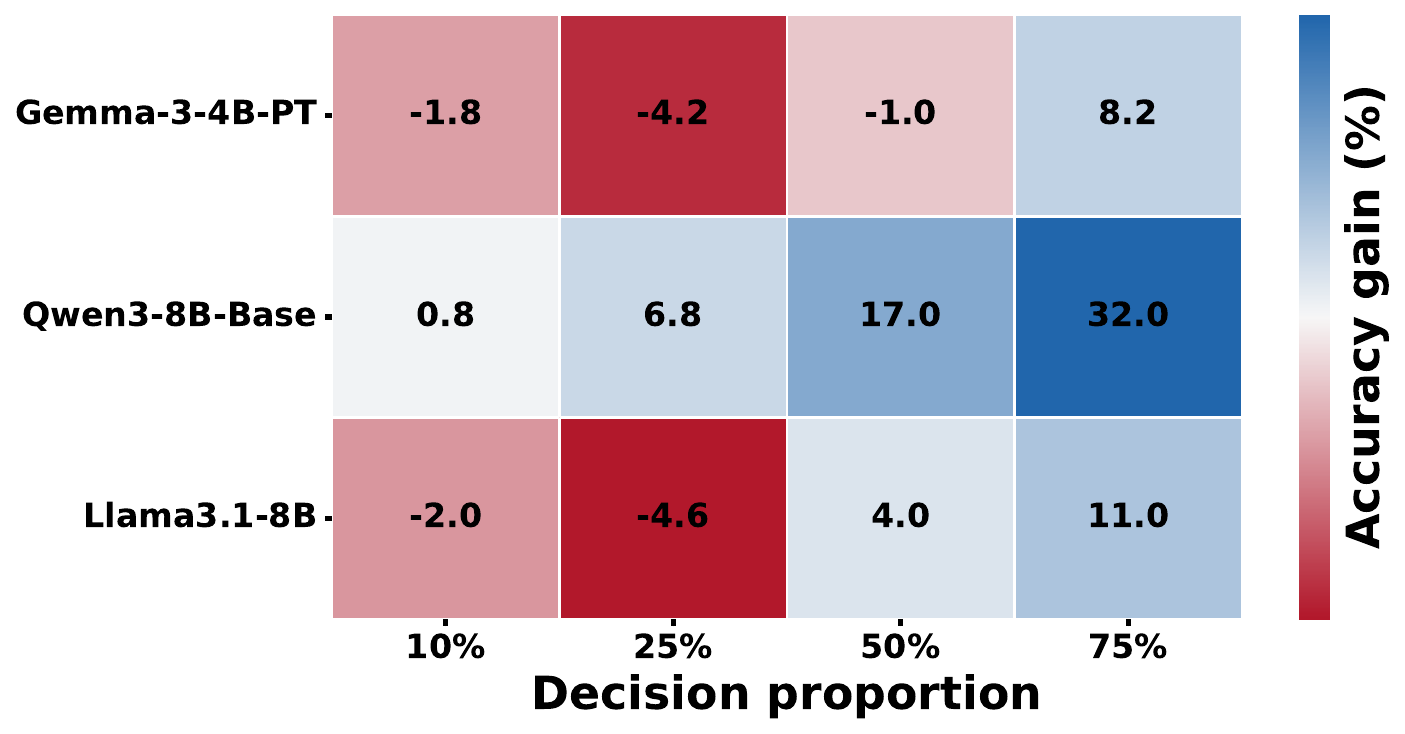}
    \caption{Effect of the verifier on \method.
    We report the accuracy gains of \method over directly injecting reasoning segments from Qwen3-32B into various generator models on MATH.
    }
    \label{fig:verifier_effect}
    \vspace{-15pt}
\end{figure}

\subsection{Verifier $\mathcal{V}$}\label{analysis:verifier}
We also investigate whether a verifier is necessary for \method.
We conduct a case study which directly injects the mentor's reasoning segments without verification in \cref{fig:verifier_effect}.
Empirically, we find that for stronger generator models such as Qwen3-8B-Base, the verifier is crucial for achieving performance gains. Without it, the model suffers substantial degradation at larger decision proportions because the mentor’s reasoning segments introduce redundant or unhelpful patterns.
In contrast, for weaker generators such as Llama3.1-8B and Gemma3-4B-PT, the verifier may cause \method to be overly conservative, reducing its willingness to adopt mentor reasoning that could otherwise be beneficial. 
Nevertheless, when the decision proportion becomes large, all models eventually exhibit performance drops due to excessive mentor intervention.
We provide more empirical results in \cref{appendix:verifier}.
Overall, these findings highlight that the verifier plays a critical role in balancing the mentor’s guidance with the generator’s autonomy.

\subsection{Segment Size $L$}
\begin{figure}[t]
    \centering
    \begin{subfigure}{\linewidth}
        \includegraphics[width=\linewidth]{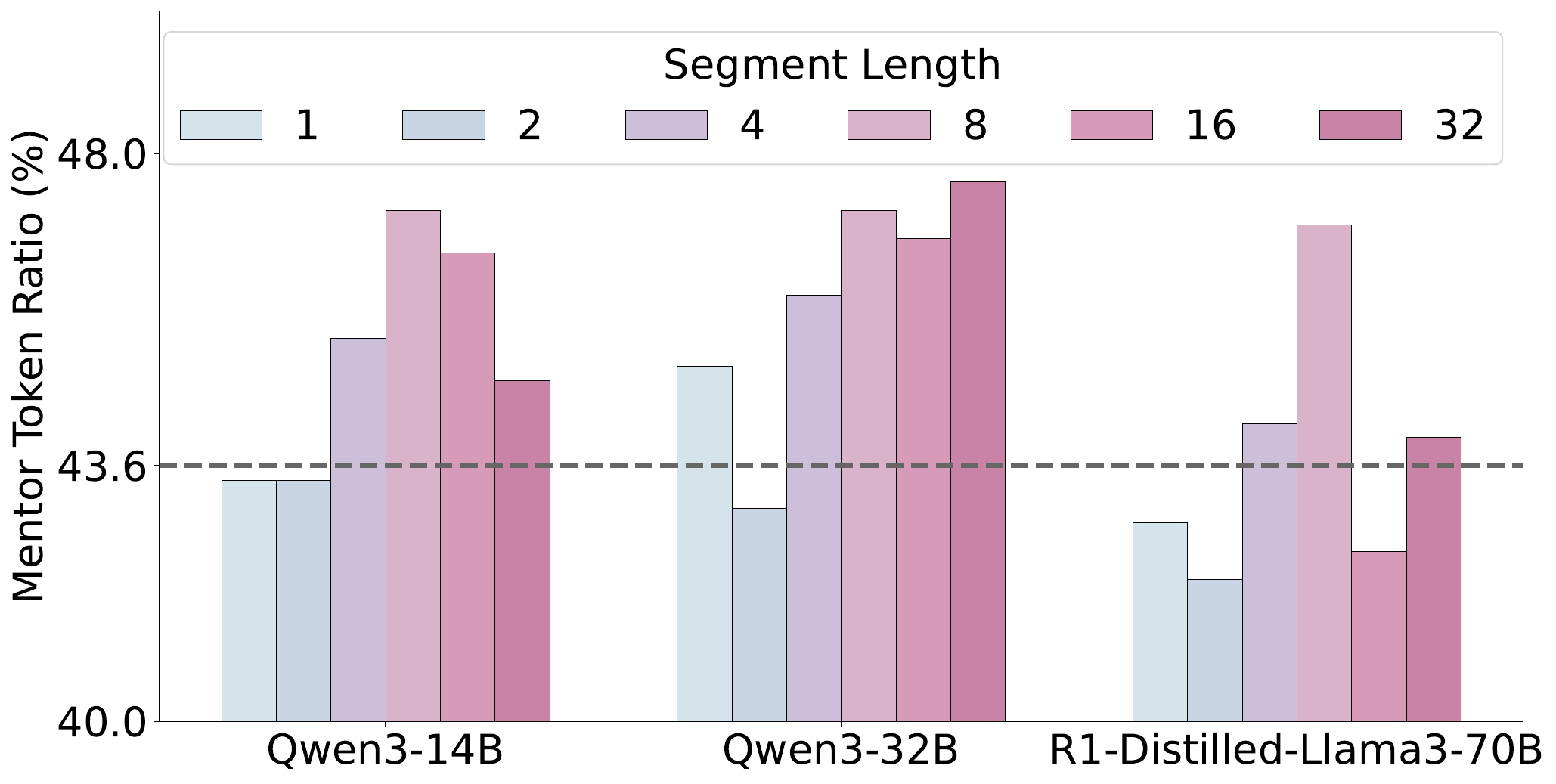}
        \caption{Accuracy ($\%$) across different segment sizes.}
    \end{subfigure}

    \vspace{6pt}

    \begin{subfigure}{\linewidth}
        \includegraphics[width=\linewidth]{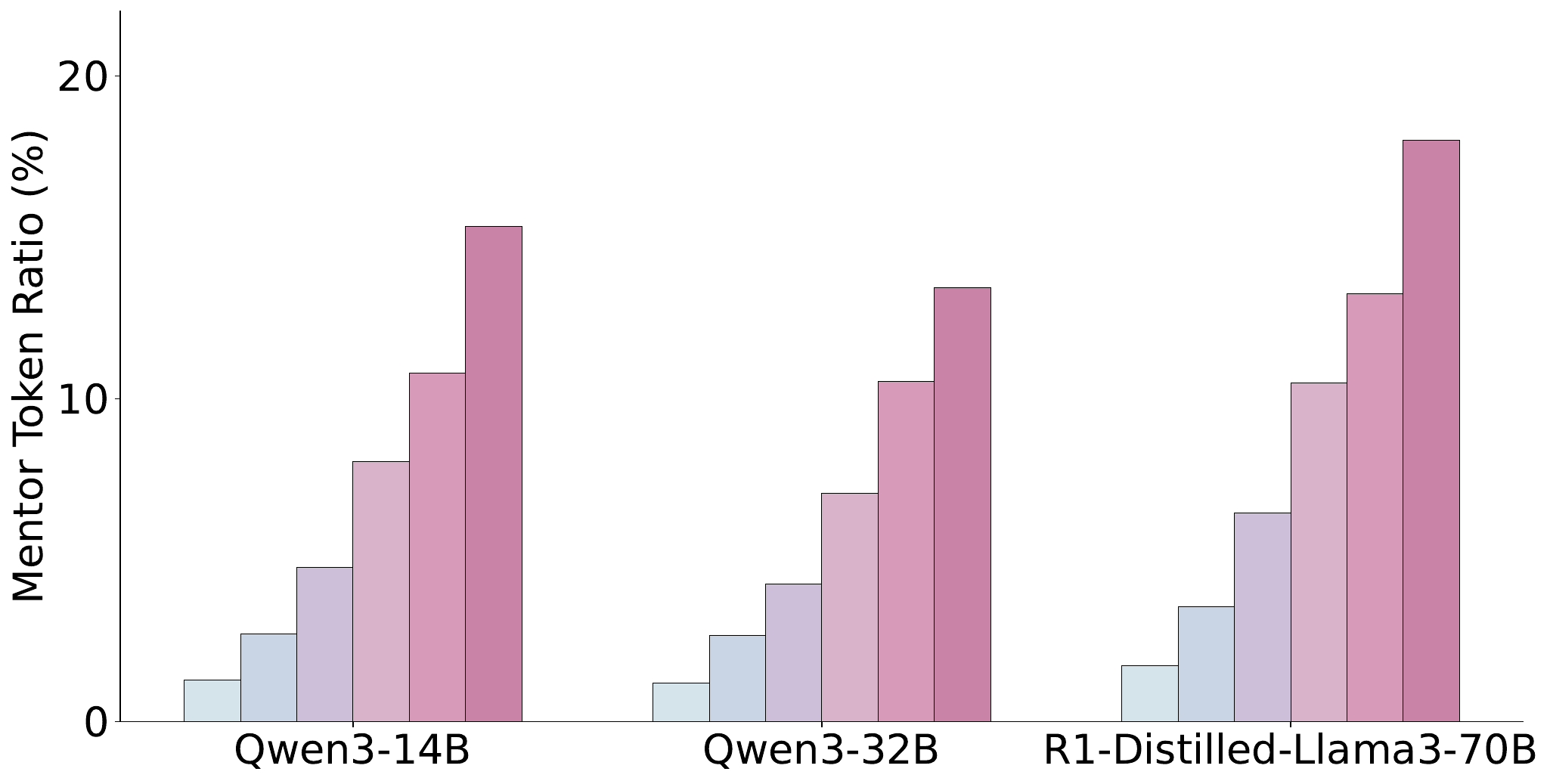}
        \caption{Mentor token ratio ($\%$) across different segment sizes.}
    \end{subfigure}
    \caption{Analysis of \methodtrain on segment size. We report the performance and mentor token ratio of 6 different segment sizes with multiple mentor LRMs on MATH for generator model Qwen3-8B-Base.
    }
    \label{fig:segment_size}
\end{figure}

We explore the impact of segment sizes in this section.
While longer segments provide the verifier with more contextual information, they also increase both computational cost and the risk of introducing redundant reasoning content.
As shown in \cref{fig:segment_size}, we observe that longer segments do not consistently lead to better performance; in several cases, moderate or even shorter segments achieve higher accuracy.
Notably, increasing the segment size often results in a substantial rise in the mentor token ratio (\cref{fig:segment_size}b), without a corresponding improvement in accuracy (\cref{fig:segment_size}a).
This suggests that the verifier benefits more from concise and focused reasoning cues than from overly long reasoning spans, which may obscure decisive signals with irrelevant details.
Overall, these results indicate that effective collaboration in \methodtrain does not require long reasoning segments; moderate segment sizes suffice to deliver accurate guidance while keeping the final output concise.

\subsection{Case Study}
\begin{figure}[t]
    \centering
    \includegraphics[width=\linewidth]{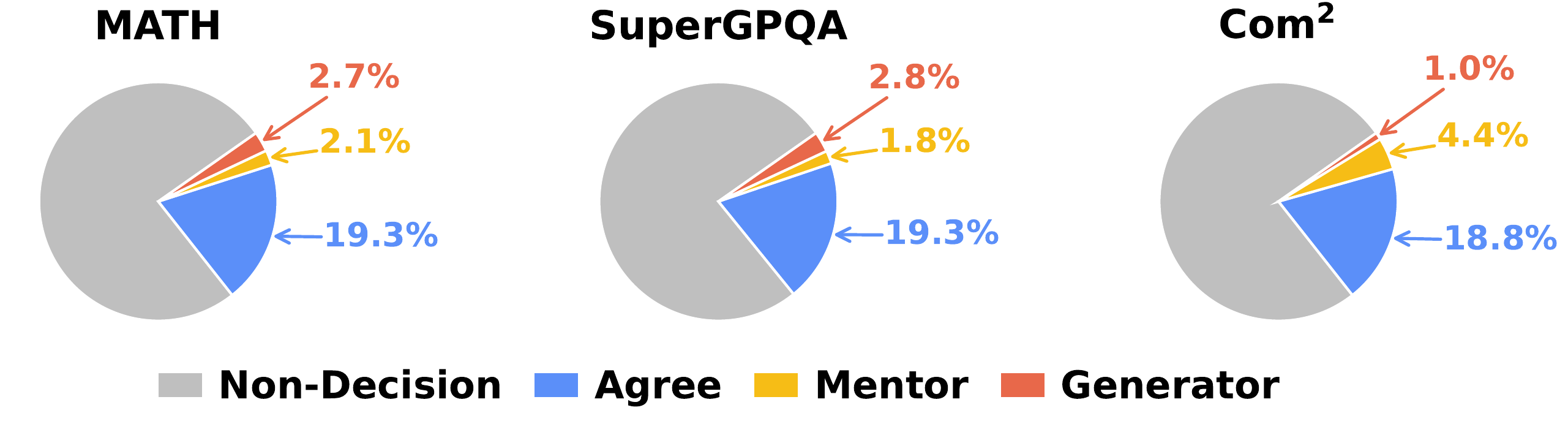}
    \caption{Token-level composition of inference under \methodtrain. We report the collaboration between Qwen3-8B-Base and Qwen3-14B.
    }
    \label{fig:case_study}
\end{figure}

\begin{table*}[t]
    \centering
    \small
    \setlength{\tabcolsep}{4pt}
    \renewcommand{\arraystretch}{1.2}
    \begin{tabularx}{\linewidth}{@{}>{\centering\arraybackslash}m{0.08\linewidth} X@{}}
    \toprule
    \multirow{3}{*}{\raisebox{-2\height}{\rotatebox{90}{\textbf{MATH}}}} &
    \textbf{Question:} If $x$, $y$, and $z$ are positive with $xy = 24$, $xz = 48$, and $yz = 72$, then find $x + y + z.$ \\[2pt]
    & \textbf{Generator:} Multiplying the three equations gives $(xyz)^2 = 24 \cdot 48 \cdot 72 = 82944$.  Taking the square root of both sides gives $xyz = 288$.  Dividing this by $yz$ gives $x = 3$.  Dividing by $xz$ gives $y = 8$.  Dividing by $xy$ gives $z = 12$.  Therefore, $x + y + z = 3 + 8 + 12 = \boxed{23}$. \\[2pt]
    & \textbf{\methodtrain with $\rho=25\%$ and segment size 16:}
    Multiplying \hlg{all three equations together gives $(xyz)^2 = 24 \cdot$} $48 \cdot 72 = 82944$.  \hlb{Taking} the square root of both sides gives $xyz = 288$.  \hlp{Dividing this by $xy = 24$ gives $z =$}$ 12$. Dividing by $xz = 48$ gives $y = 6$.\hlb{  }Dividing by \hlb{$yz$}$ = 72$ \hlb{gives} $x = 4$.  Therefore, \hlb{$x$}$ + y $\hlb{$+$}$ z = $\hlb{$4 +$}$ 6 $\hlb{$+$}$ 12 = \boxed{22}$.\\
    \midrule
    \multirow{3}{*}{\raisebox{-2\height}{\rotatebox{90}{\textbf{SuperGPQA}}}} &
    \textbf{Question:} Which of the following statements about dance are incorrect?
    1. The dance "Swan Song" belongs to contemporary ballet;
    2. The story of Swan Lake is based on a Danish fairy tale;
    3. The choreography method used in "Swan Lake Act II" is symphonic choreography;
    4. Classical ballet positions do not require symmetry.\newline
    (A) 2,3,4 (B) 2,4 (C) 1,2 (D) 2,3 (E) 1,3,4 (F) 1,4 (G) 1,3 (H) 3,4 (I) 1,2,3,4 (J) 1,2,4\\
    &\textbf{Generator:} 1. The dance "Swan Song" is indeed a contemporary ballet, so this statement is correct.2. The story of Swan Lake is based on a German fairy tale, not a Danish one, so this statement is incorrect.3. The choreography method used in "Swan Lake Act II" is not symphonic choreography, but rather a combination of classical ballet and narrative elements, so this statement is incorrect.4. Classical ballet positions do require symmetry, so this statement is incorrect.Therefore, the incorrect statements are 2, 3, and 4.\newline Answer: A. \\
    & \textbf{\methodtrain with $\rho=25\%$ and segment size 16:} 1. \hlg{"Swan Song" is not a known ballet, but "Swan Lake"} is \hlb{a} famous ballet, so this statement \hlb{is} incorrect.2. The story \hlb{of} "Swan \hlb{Lake"} is based on \hlb{a} German fairy tale, \hlb{not} a Danish \hlb{one, so this} statement \hlb{is} incorrect.3. \hlb{"Swan} Lake Act II" does not use \hlb{symphonic} choreography, so this statement is incorrect.4. Classical ballet positions \hlb{do} require symmetry, so this statement \hlb{is} incorrect.Therefore, all statements are incorrect.\newline Answer: I. \\

    \bottomrule
    \end{tabularx}
    
    \caption{Two case studies of \methodtrain on MATH and SuperGPQA. By selectively injecting a few tokens with the help of the mentor and verifier, \methodtrain guides the generator to concise and correct answers. We highlight tokens that both models agree on in \hlb{blue}, tokens that the generator maintains after verification in \hlp{red}, and tokens that the mentor inspires the generator in \hlg{green}.}
    \label{tab:mentorcollab_case}
    \end{table*}

We analyze when tokens trigger the verifier and when mentor reasoning is adopted under the \methodtrain inference mechanism.
As shown in \cref{fig:case_study}, only a small fraction of generation positions (approximately 5\%) trigger the verifier for explicit decision making, and among them, roughly 2\% further receive mentor-provided inspirations.
This indicates that \methodtrain applies collaboration in a sparse and targeted manner, rather than relying on frequent external guidance throughout generation.
To better understand how these sparse interventions influence generation behavior, we present a qualitative case study in \cref{tab:mentorcollab_case}.
As shown in the table, the vast majority of tokens remain unchanged or none-investigated across the two models, while only a few positions trigger verifier decisions and mentor inspirations.
Despite their small quantity, these mentor-injected tokens occur at critical reasoning steps and are sufficient to correct erroneous intermediate conclusions, ultimately leading the generator to the correct final answer.
\cref{app:case_study} further demonstrate the effectiveness of \method in correcting reasoning errors through targeted mentor interventions.






\section{Related Work}

\paragraph{Routing-Based Model Collaboration.}
These strategies route queries to the most cost-efficient and high-performing model within a diverse pool of LLMs \cite{feng2025one}. 
This line of work includes two major variants. 
The first focuses on selecting the most appropriate model for a given query before any inference takes place. 
Representative methods include non-parametric routers \citep{shnitzer2023large}, parametric routers \citep{jiang-etal-2023-llm, hu2024routerbench, ding2024hybrid, ong2025routellm}, domain-specific routers \citep{lu-etal-2024-routing, stripelis-etal-2024-tensoropera}, and graph-based routing frameworks \citep{feng2025graphrouter}.
The second variant invokes additional models only when the current model produces unsatisfactory outputs \interalia{gupta2024language, yue2024large, chen2024frugalgpt}.
A recent line routes at the token level, invoking the larger model only at divergence points it deems outcome-changing~\citep{fu2025rr, yang2025speculative}; unlike \method, the generator there adopts the mentor's continuation rather than verifying and possibly rejecting it.
In contrast, \method uses two models to collaboratively generate a single sequence, substantially reducing computational cost while enabling more fine-grained cooperation.

\paragraph{Fusion-Based Model Collaboration.}
Fusion-based methods fall into two categories: token-level distribution fusion and criterion-based model switching during generation.
Fusing distributions combines multiple models by jointly contributing to each next-token prediction, typically through logit-level operations such as contrasting weights across models or layers \interalia{liu-etal-2021-dexperts, li-etal-2023-contrastive, liu2024tuning, chuang2024dola, mavromatis2024pack, DBLP:journals/corr/abs-2506-15710}, which requires low-level access to model internals.
Switching methods train a router or rely on heuristics to decide which model to follow at each token position, either via rule-based criteria \citep{fei-etal-2025-nudging, wang2025speculate, chen2025r} or learning-based mechanisms \citep{shen-etal-2024-learning, zheng2025citer, feng2025don}.
\method selectively adopts mentor LRM reasoning only when helpful, allowing the generator SLM to lead a concise yet effective reasoning process.
\section{Conclusion}
We propose \method, an inference-time model collaboration framework that treats LRMs as mentors and SLMs as generators. 
\method steers the SLM only when guidance is likely to help, preserving the generator's concise reasoning style via lightweight, reusable verifiers that bypass computationally intensive joint optimization. 
Across diverse generator–mentor pairs, \method yields consistent improvements of 3–8 absolute points while introducing as few as 10\% of mentor tokens under a short-token budget, making it suitable for scenarios where concise reasoning is desired. 
Further analyses show \method is robust to its key hyperparameters and generalizes across domains and models.
We expect future work to build upon \method by developing more efficient verifiers, advancing scalable and robust model collaborations.
\section*{Limitations}
Our study could be extended in the following ways. 
First, \method largely relies on random positions to initiate mentor collaboration. 
A more precise criterion for detecting intervention needs would further decrease the computational overhead associated with dual-model inference.
Second, it remains unclear why high-likelihood generator predictions are bypassed in certain contexts, warranting future analysis of internal representations to uncover the underlying causes.
Finally, our study focuses on a dual-model setting with a single generator and mentor, and extending \method to multi-mentor collaboration setup is left as future work.
We hope this work motivates further exploration of more fine-grained and computationally efficient paradigms for model collaborations.

\section*{Ethics Statement}
We use only public benchmarks and do not collect user data; nonetheless, the approach inherits biases of underlying LMs and may amplify mentor biases through selective segment adoption, and it increases data exposure at deployment due to multi-model querying; thus any real-world use should include bias audits, content filtering, and strict data-retention controls.

\section*{Acknowledgements}
This research was developed with funding from the Defense Advanced Research Projects Agency's (DARPA) SciFy program (Agreement No. HR00112520300) and NSF Grant No. IIS2142739. The views expressed are those of the author and do not reflect the official policy or position of the Department of Defense or the U.S. Government.
\bibliography{reference}
\newpage
\appendix
\section{Additional Experimental Results} 
\subsection{Average Performance of \method} \label{app:average}
\begin{figure}[t]
    \includegraphics[width=\linewidth]{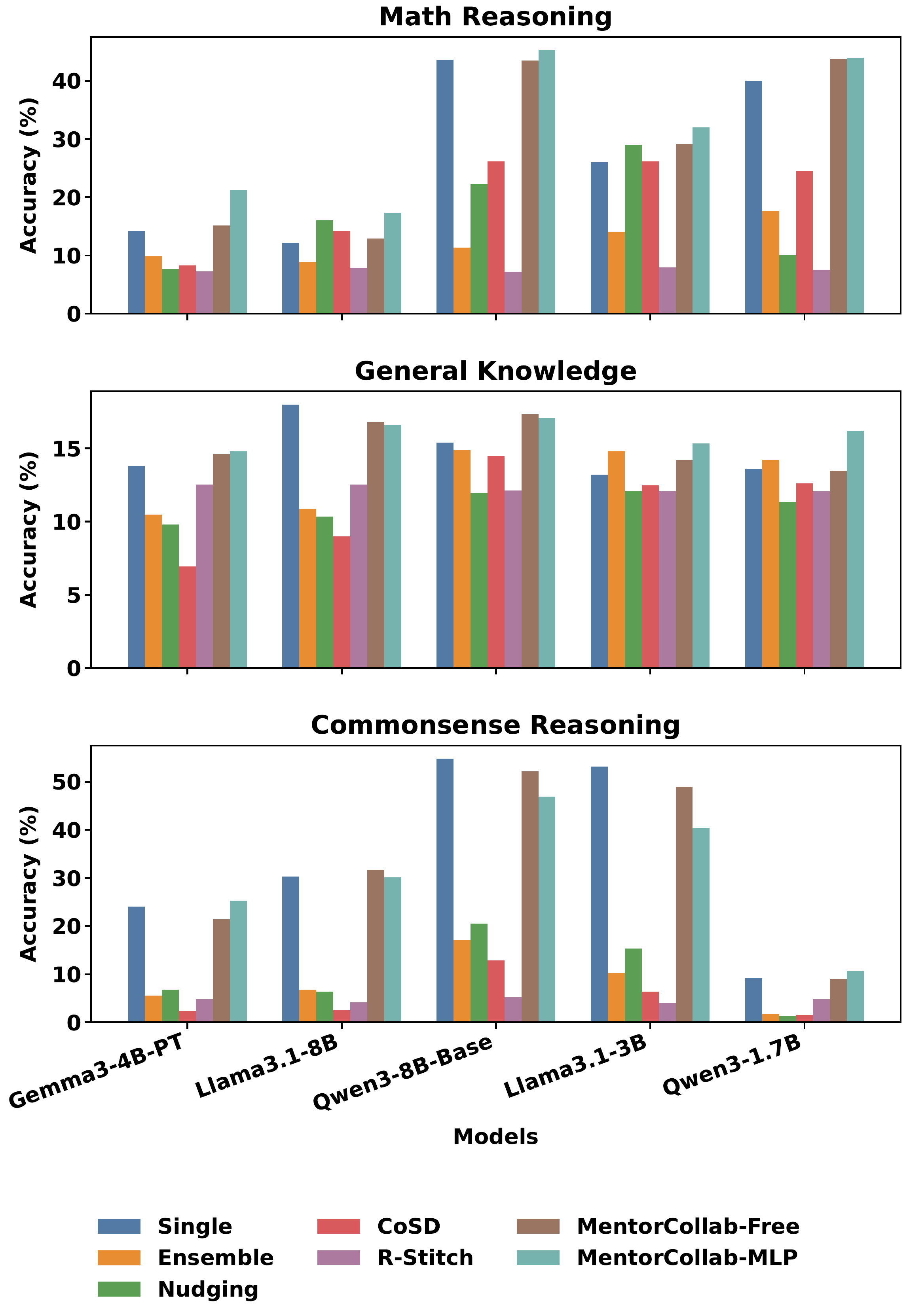}
    \vspace{-15pt}
    \caption{
        Average performance of \method across different generator–mentor model pairs.
Beyond significantly outperforming baseline collaboration methods, \method improves generator performance in 12 out of 15 model–domain combinations, with gains of up to 7 percentage points on average.
    }
    \vspace{-5pt}
    \label{fig:average}
\end{figure}

We provide the average performance of \method in \cref{fig:average}.
Empirically, we find \method achieves better performance in 5 out of 5 generator models on math reasoning, 4 out of 5 generator models on general knowledge, and 3 out of 5 generator models on commonsense reasoning.
Performance gains exhibit an inverse correlation with the generator's intrinsic capability. 
For weaker models (e.g., Qwen3-1.7B), the mentor acts as a critical crutch, correcting fundamental reasoning flaws.
However, for stronger baselines (e.g., Qwen3-8B-Base in Commonsense tasks), the generator is already approaching the domain ceiling, where external intervention becomes a source of noise rather than signal.
In these high-competence regimes, the challenge shifts from providing guidance to preventing interference.
Unlike previous collaboration methods (Nudging, CoSD, R-Stitch) which exhibit high volatility—often degrading performance significantly below the single model baseline—\method demonstrates superior stability.
Our verification mechanism effectively acts as a safety gate, filtering out harmful interventions that plague other methods.

\subsection{Detailed Analysis of Verifier Effectiveness}\label{appendix:verifier}
\begin{figure}[t]
    \includegraphics[width=\linewidth]{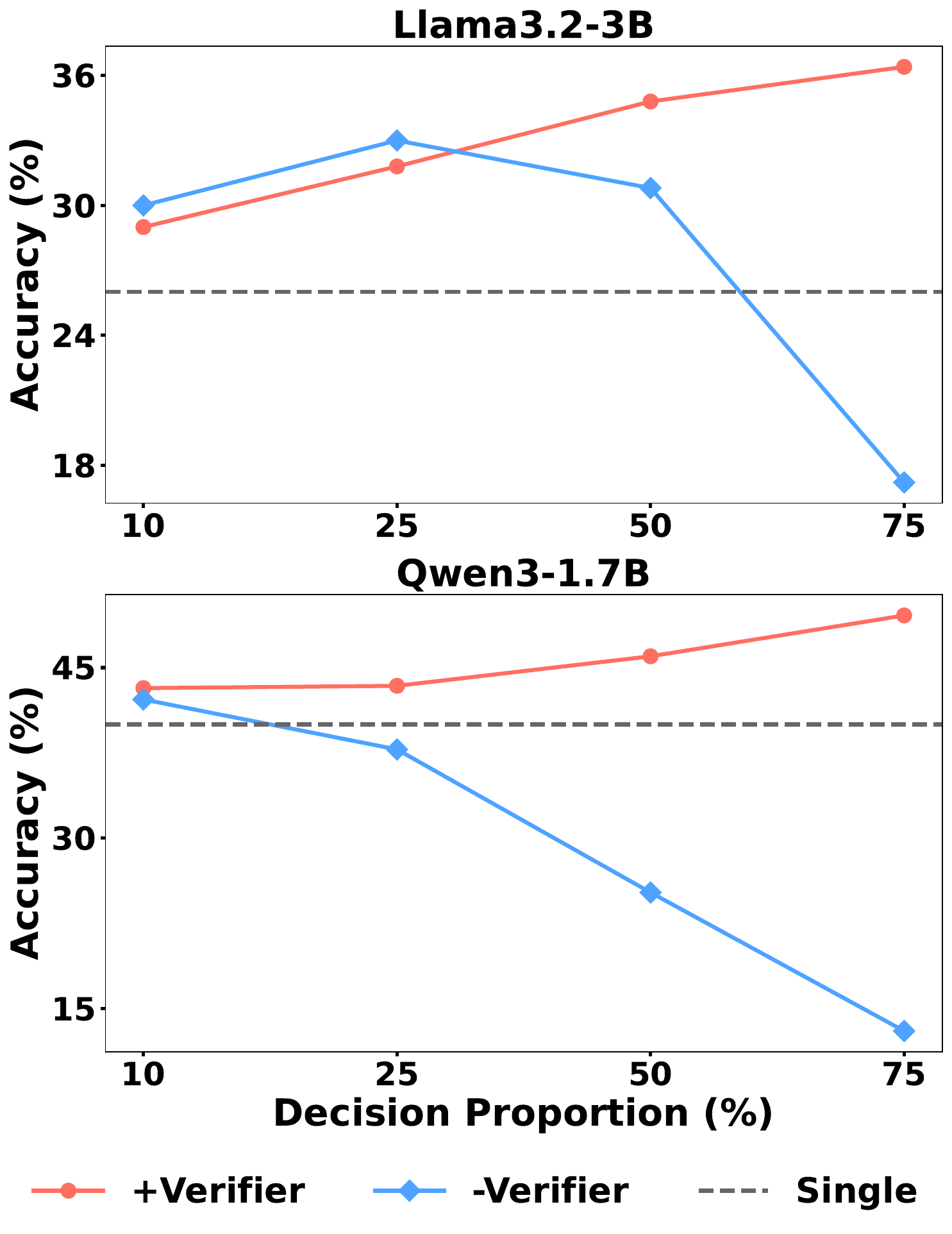}
    \caption{Detailed analysis of the verifier effectiveness.
    We report the accuracy gains of \method over directly injecting reasoning segments from Qwen3-14B into two more generator models on MATH.
    }
    \vspace{-5pt}
    \label{fig:verifier_effectiveness}
\end{figure}

To further investigate the role of the verifier module in \methodtrain, we conduct an ablation study comparing our method against direct injection in \cref{analysis:verifier}.
As shown in \cref{fig:verifier_effectiveness}, we further analyze the role of the verifier by comparing \methodtrain with direct injection on two more generator models on MATH.
Across both generator models, \methodtrain consistently benefits from incorporating the verifier, whereas direct injection without verification exhibits increasingly unstable behavior as the decision proportion grows.
Specifically, the verifier enables the generator to selectively accept helpful mentor-provided reasoning while rejecting ineffective or misaligned segments, thereby improving its own reasoning process.
In contrast, when reasoning segments are injected without verification, the generator tends to indiscriminately follow the mentor’s reasoning patterns as the level of intervention increases, which can overwhelm the generator’s internal reasoning and lead to performance degradation, resembling the overthinking behavior of the mentor model itself.

\subsection{Extended Evaluation on Training-Based Methods}\label{app:training-based}

We compare \methodtrain with Co-LLM~\citep{shen-etal-2024-learning}, a training-based collaboration method, on the MATH domain.
We evaluate on three generator–mentor pairs with compatible tokenizers, as Co-LLM requires tokenizer compatibility and supervised training on ground-truth sequences.

\begin{table*}[t]
  \centering
  \small
  \begin{tabular}{cccccc}
  \toprule
  \textbf{\makecell{Generator\\Name}} & \textbf{\makecell{Mentor\\Name}} & \textbf{Generator} & \textbf{Mentor} & \textbf{Co-LLM} & \textbf{\makecell{\method\\\textsc{MLP}}} \\
  \midrule
  Qwen3-8B-Base & Qwen3-14B & 43.60\% & 2.80\% & 25.00\% & \textbf{46.60\%} \\
  Qwen3-8B-Base & Qwen3-32B & 43.60\% & 3.20\% & 21.80\% & \textbf{46.80\%} \\
  Llama3.1-8B & R1-70B & 12.20\% & 21.00\% & 4.40\% & \textbf{18.00\%} \\
  \bottomrule
  \end{tabular}
  \caption{Training-based comparison between Co-LLM~\citep{shen-etal-2024-learning} and \methodtrain on MATH. \methodtrain achieves consistent performance gains without relying on expensive collaboration-specific training.}
  \label{tab:training-comparison}
\end{table*}

\section{Comparison with Speculative Decoding}\label{app:spec_decoding}

A natural question is whether Speculative Decoding~\citep{DBLP:conf/icml/LeviathanKM23} should serve as a baseline for \method. We argue that the two address fundamentally different objectives, despite both involving an SLM--LRM pair.

\paragraph{Different goals.} Speculative Decoding uses the SLM to \emph{draft for the LRM}: the SLM proposes candidate tokens that the LRM verifies, aiming to reproduce the LRM’s output faster without changing its distribution. In contrast, \method uses the LRM to \emph{inspire the SLM}: the mentor provides concise guidance at critical points to make the generator more accurate, while keeping the generator in control of decoding. In other words, Speculative Decoding accelerates the large model, whereas \method improves the small model.

\paragraph{Different assumptions.} Our setting assumes the LRM cannot reliably solve the task within the token budget---mentor LRMs frequently collapse mid-reasoning under greedy decoding, consistent with prior observations~\citep{sui2025stop}. Speculative Decoding, by contrast, assumes the LRM is the \emph{target} model whose output is worth preserving exactly.

\paragraph{Why not as a baseline?} Since Speculative Decoding preserves the LRM’s output distribution by design, it would simply reproduce the LRM’s performance in our evaluation---which, as shown in our main results, is often poor due to overthinking and mid-reasoning collapse under constrained token budgets. The efficiency of \method lies not in reproducing verbose LRM reasoning faster, but in \emph{producing more concise yet accurate reasoning chains} by selectively incorporating only helpful mentor segments.

\section{\methodtrain Training}\label{app:verifier_training}
\subsection{Training Data}
We train two versions of \methodtrain, one on the training split of MATH\citep{hendrycksmath2021} for the mathematical reasoning domain, and one on the training split of SuperGPQA\citep{DBLP:journals/corr/abs-2502-14739} for the general knowledge domain.
Empirically, we find that Qwen3-8B-Base \citep{DBLP:journals/corr/abs-2505-09388} performs stably in the setting of \methodfree, and we use Qwen3-14B\citep{DBLP:journals/corr/abs-2505-09388} as the mentor models under this setting for sampling of the training data.
We perform \methodfree on the whole training split of MATH, and randomly sample 4000 examples from the training split of SuperGPQA for the general knowledge domain.

Empirically, we find that the sampled data is not balanced, where the distribution of labels---corresponding to whether the verifier selects the generator's segment or the mentor's segment---is skewed. 
Such imbalance may cause the verifier to learn trivial class priors rather than meaningful reasoning discrepancies. 
To address this, we apply a balancing strategy by randomly downsampling the majority class to achieve an approximately 1:1 ratio between samples where the generator is preferred and those where the mentor is preferred.
\subsection{Hyperparameters}
The verifier is implemented as a lightweight MLP that takes the generator’s hidden state at the decision point as input and predicts which branch to follow.
Specifically, the MLP consists of three hidden layers with widths $\{2d, d, d/2\}$, where $d$ denotes the hidden size of the generator model.
Each hidden layer is followed by Batch Normalization and ReLU activation, with a dropout rate of 0.1 applied throughout.
The final layer is a linear projection to a scalar, followed by a sigmoid function to produce a probability score.
During inference, scores above 0.5 indicate selecting the mentor-guided branch, while lower scores correspond to retaining the generator’s own continuation.

\section{Prompt and Evaluation} \label{app:prompts}
\subsection{MATH}
We provide a four-shot prompt template below for MATH.
For evaluation, we extract the \verb|\boxed{}| answers from the model's responses, comparing them with the ground-truth answers, and calculate the accuracy.
\begin{promptbox}{MATH}
    Think step by step.

    \vspace{0.5em}
    
    \textbf{Problem:} How many vertical asymptotes does the graph of $y=\frac{2}{x^2+x-6}$ have?
    
    \textbf{Solution:} The denominator of the rational function factors into $x^2+x-6=(x-2)(x+3)$. Since the numerator is always nonzero, there is a vertical asymptote whenever the denominator is $0$, which occurs for $x = 2$ and $x = -3$. Therefore, the graph has $\boxed{2}$ vertical asymptotes.

    \vspace{0.5em}

    \textbf{Problem:} If $5x - 3 = 12$, what is the value of $5x + 3$?
    
    \textbf{Solution:} Adding 6 to both sides of $5x - 3 =12$ gives $5x -3 + 6 = 12 + 6$. Simplifying both sides gives $5x + 3 = \boxed{18}$.

    \vspace{0.5em}

    \textbf{Problem:} Ryan has 3 red lava lamps and 3 blue lava lamps. He arranges them in a row on a shelf randomly, then turns 3 random lamps on. What is the probability that the leftmost lamp on the shelf is red, and the leftmost lamp which is turned on is also red?
    
    \textbf{Solution:} There are $\binom{6}{3}=20$ ways for Ryan to arrange the lamps, and $\binom{6}{3}=20$ ways for him to choose which lamps are on, giving $20\cdot20=400$ total possible outcomes. There are two cases for the desired outcomes: either the left lamp is on, or it isn't. If the left lamp is on, there are $\binom{5}{2}=10$ ways to choose which other lamps are on, and $\binom{5}{2}=10$ ways to choose which other lamps are red. This gives $10\cdot10=100$ possibilities. If the first lamp isn't on, there are $\binom{5}{3}=10$ ways to choose which lamps are on, and since both the leftmost lamp and the leftmost lit lamp must be red, there are $\binom{4}{1}=4$ ways to choose which other lamp is red. This case gives 40 valid possibilities, for a total of 140 valid arrangements out of 400. Therefore, the probability is $\dfrac{140}{400}=\boxed{\dfrac{7}{20}}$.

    \vspace{0.5em}

    \textbf{Problem:} Alice wants to buy $3$ pounds of veal at the grocery store, but the scales at the store only show weight in kilograms. If one kilogram is $2.20$ pounds, how many kilograms of veal should Alice buy? (You may use a calculator on this problem; answer to the nearest hundredth.)
    
    \textbf{Solution:} Since Alice wants to buy $3$ pounds of veal, we multiply the quantity of $3$ pounds by the conversion factor $\frac{1\ \text{kg}}{2.20\ \text{lb}}$ to obtain $3\ \text{lb} \cdot \frac{1\ \text{kg}}{2.20\ \text{lb}} \approx \boxed{1.36}\ \text{kg}$.

    \vspace{0.5em}

    \textbf{Problem:} \texttt{<Problem>}

    \textbf{Solution:}
\end{promptbox}

\subsection{SuperGPQA} \label{app:supergpqa}
We provide a five-shot prompt template below for SuperGPQA, following the prompt template in \citet{DBLP:journals/corr/abs-2502-14739}.
We use the evaluation protocol following \citet{DBLP:conf/nips/WangMZNCGRAHJLK24}.
Specifically, we employ a cascaded extraction strategy: we initially attempt to parse the answer using the regular expression \verb|'answer is \(?\([A-J]\)?\)'|, fall back to a secondary pattern \verb|'\.*\[aA\]nswer:\s*\([A-J]\)'|for formatting deviations, and default to wrong answer if both methods fail.
\begin{promptbox}{SuperGPQA}
    You are an expert in science who answers multiple choice questions step by step.
    
    Answer the following multiple choice question. There is only one correct answer. The last line of your response should be in the format 'Answer: \$LETTER' (without quotes), where LETTER is one of A, B, C, D, E, F, G, H, I, or J.

    \vspace{1em}

    \textbf{Question:} 
    A refracting telescope consists of two converging lenses separated by 100 cm. The eye-piece lens has a focal length of 20 cm. The angular magnification of the telescope is\\
    A) 10\\
    B) 40\\
    C) 6\\
    D) 25\\
    E) 15\\
    F) 50\\
    G) 30\\
    H) 4\\
    I) 5\\
    J) 20

    \textbf{Answer:} Let's think step by step. In a refracting telescope, if both lenses are converging, the focus of both lenses must be between the two lenses, and thus the focal lengths of the two lenses must add up to their separation. Since the focal length of one lens is 20 cm, the focal length of the other must be 80 cm. The magnification is the ratio of these two focal lengths, or 4.\\
    Answer: H.

    \vspace{1em}

    \textbf{Question:} 
    Say the pupil of your eye has a diameter of 5 mm and you have a telescope with an aperture of 50 cm. How much more light can the telescope gather than your eye? \\
    A) 1000 times more\\
    B) 50 times more\\
    C) 5000 times more\\
    D) 500 times more\\
    E) 10000 times more\\
    F) 20000 times more\\
    G) 2000 times more\\
    H) 100 times more\\
    I) 10 times more\\
    J) N/A

    \textbf{Answer:} Let's think step by step. The amount of light a telescope can gather compared to the human eye is proportional to the area of its apertures. The area of a circle is given by the formula $A = \pi \left(\frac{D}{2}\right)^2$, where $D$ is the diameter. Therefore, the relative light-gathering power is calculated as:
    \[
    \frac{\left(\frac{50 \text{ cm}}{2}\right)^2}{\left(\frac{5 \text{ mm}}{2}\right)^2} = \frac{\left(\frac{50 \text{ cm}}{0.1 \text{ cm}}\right)^2}{\left(\frac{5 \text{ mm}}{0.1 \text{ cm}}\right)^2} = \frac{500^2}{5^2} = 10000.
    \]
    Answer: E.

    \vspace{1em}

    \textbf{Question:} 
    Where do most short-period comets come from and how do we know? \\
    A) The Kuiper belt; short period comets tend to be in the plane of the solar system like the Kuiper belt. \\
    B) The asteroid belt; short period comets tend to come from random directions indicating a spherical distribution of comets called the asteroid belt. \\
    C) The asteroid belt; short period comets tend to be in the plane of the solar system just like the asteroid belt. \\
    D) The Oort cloud; short period comets have orbital periods similar to asteroids like Vesta and are found in the plane of the solar system just like the Oort cloud. \\
    E) The Oort Cloud; short period comets tend to come from random directions indicating a spherical distribution of comets called the Oort Cloud. \\
    F) The Oort cloud; short period comets tend to be in the plane of the solar system just like the Oort cloud. \\
    G) The asteroid belt; short period comets have orbital periods similar to asteroids like Vesta and are found in the plane of the solar system just like the asteroid belt. 

    \textbf{Answer:} Let's think step by step. Most short-period comets originate from the Kuiper belt. This is deduced from the observation that these comets tend to follow orbits that lie in the plane of the solar system, similar to the distribution of objects in the Kuiper belt itself. Thus, the alignment of these cometary orbits with the ecliptic plane points to their Kuiper belt origin.\\
    Answer: A.

    \vspace{1em}

    \textbf{Question:} 
    Colors in a soap bubble result from light \\
    A) dispersion \\
    B) deflection \\
    C) refraction \\
    D) reflection \\
    E) interference \\
    F) converted to a different frequency \\
    G) polarization \\
    H) absorption \\
    I) diffraction \\
    J) transmission 

    \textbf{Answer:} Let's think step by step. The colorful patterns observed in a soap bubble are caused by the phenomenon of light interference. This occurs when light waves bounce between the two surfaces of the soap film, combining constructively or destructively based on their phase differences and the varying thickness of the film. These interactions result in vibrant color patterns due to variations in the intensity of different wavelengths of light.\\
    Answer: E.

    \vspace{1em}

    \textbf{Question:} 
    A microwave oven is connected to an outlet, 120 V, and draws a current of 2 amps. At what rate is energy being used by the microwave oven? \\
    A) 240 W\\
    B) 120 W\\
    C) 10 W\\
    D) 480 W\\
    E) 360 W\\
    F) 200 W\\
    G) 30 W\\
    H) 150 W\\
    I) 60 W\\
    J) 300 W

    \textbf{Answer:} Let's think step by step. The rate of energy usage, known as power, in an electrical circuit is calculated by the product of voltage and current. For a microwave oven connected to a 120 V outlet and drawing a current of 2 amps, the power consumption can be calculated as follows:
    \[
    \begin{aligned}
    \text{Power} &= \text{Voltage} \times \text{Current} \\
                 &= 120 \, \text{V} \times 2 \, \text{A} = 240 \, \text{W}.
    \end{aligned}
    \]
    Therefore, the microwave oven uses energy at a rate of 240 watts.\\
    Answer: A.

    \vspace{1em}

    \textbf{Question:} \texttt{<Question>}

    \textbf{Answer:}
\end{promptbox}
\subsection{Com$^{2}$-Hard Intervention}
We provide the zero-shot prompt template below for Com$^{2}$-Hard Intervention, as used in \citet{xiong-etal-2025-com2}.
The evaluation protocol is the same as in \cref{app:supergpqa}.
\begin{promptbox}{Com\textsuperscript{2}-Hard Intervention}
    You are given a puzzle about a crime, which include a crime statement, a set of facts, a question, and a set of suspects (options). You should conduct detailed analysis to find the most plausible suspect as the criminal. At the end of your answer, choose answer to the question in format such as \verb|'Answer: A)____.'|

    \vspace{1em}

    \textbf{Crime:} 
    \texttt{<Crime>}
    \vspace{1em}

    \textbf{Facts:}
    \texttt{<Facts>}
    \vspace{1em}

    \textbf{Question:}
    \texttt{<Question>}
    \vspace{1em}

    \textbf{Options:}
    \texttt{<Options>}
    \vspace{1em}

    \textbf{Detailed analysis:}
\end{promptbox}

\section{Case Study}\label{app:case_study}
We provide several case studies here to showcase the effectiveness of \method.
We compare single-model baselines (generator-only and mentor-only) and \method.
In \cref{tab:case_study_1}, while the mentor ultimately arrives at the correct answer, it continues to engage in extensive self-questioning and verification, resulting in a lengthy reasoning process.
In contrast, the generator requires only a short reasoning span once properly guided.
\methodtrain selectively extracts concise mentor guidance to correct the generator’s initial error, achieving the correct answer without inheriting the mentor’s overextended reasoning trajectory.
Notably, CoSD \citep{wang2025speculate} interventions often introduce hesitation tokens (e.g., “wait”), which can trigger repeated self-correction and unnecessary self-doubt, leading to prolonged and redundant reasoning traces.
In \cref{tab:case_study_2}, although CoSD introduces mentor-generated tokens into the decoding process, these interventions fail to provide an effective corrective signal.
As a result, the generator continues to commit to its original (incorrect) answer, despite being exposed to additional mentor content.
In contrast, \method enables structured collaboration that selectively incorporates mentor guidance only when it is crucial, allowing the generator to revise its answer and recover the correct prediction.
Overall, these case studies highlight that effective inference-time collaboration hinges on selective, decision-relevant mentor intervention, rather than the indiscriminate injection of additional reasoning tokens.

\begin{table*}[t]
    \centering
    \small
    \setlength{\tabcolsep}{4pt}
    \renewcommand{\arraystretch}{0}
    \begin{tabularx}{\linewidth}{@{}>{\centering\arraybackslash}m{0\linewidth} X@{}}
    \toprule
    \multirow{5}{*} &
    \textbf{Question:} Let $f$ be the function defined by $f(x) = x^3 - 49x^2 + 623x - 2015,$ and let $g(x) = f(x + 5).$  Compute the sum of the roots of $g.$
    \\ \midrule
    & \textbf{Generator:} First, we note that $g(x) = f(x + 5) = (x + 5)^3 - 49(x + 5)^2 + 623(x + 5) - 2015.$  We can expand this expression to get $g(x) = x^3 + 15x^2 + 75x + 125 - 49x^2 - 490x - 1225 + 623x + 3115 - 2015.$  Simplifying, we get $g(x) = x^3 - 15x^2 + 108x + 75.$  We can use Vieta's formulas to find the sum of the roots of $g(x).$  Vieta's formulas tell us that the sum of the roots of a polynomial is the opposite of the coefficient of the $x^{n - 1}$ term, where $n$ is the degree of the polynomial.  In this case, the degree of the polynomial is $3,$ so the sum of the roots of $g(x)$ is the opposite of the coefficient of the $x^2$ term, which is $-15.$  Therefore, the sum of the roots of $g(x)$ is $\boxed{-15}.$ \\ \midrule
    & \textbf{Mentor:} Okay, so I need to find the sum of the roots of the function g(x), which is defined as f(x + 5), where f(x) is given by $x^3 - 49x^2 + 623x - 2015$. Hmm, let me think about how to approach this.\newline First, I remember that for a polynomial, the sum of its roots can be found using Vieta's formulas. For a cubic equation $ax^3 + bx^2 + cx + d = 0$, the sum of the roots is $-b/a$. So if I can figure out the coefficients of g(x), then I can apply Vieta's formula directly.\newline 
    But maybe there's a smarter way without expanding the entire polynomial. Since g(x) is f(x + 5), that means it's a horizontal shift of the original function f(x). However, shifting a function horizontally doesn't change the sum of its roots. Wait, is that true? Let me think. If you shift a function left or right, the roots move accordingly, but their sum might change? Or does it?\newline 
    Wait, no. Let me consider an example. Suppose f(x) has roots r1, r2, r3. Then g(x) = f(x + 5) would have roots at x such that x + 5 = r1, r2, r3. 
    Therefore, the roots of g(x) are r1 - 5, r2 - 5, r3 - 5. Therefore, the sum of the roots of g(x) would be (r1 + r2 + r3) - 15. 
    So if I can find the sum of the roots of f(x), then subtract 15, that would give me the sum of the roots of g(x).\newline
    Alternatively, if I use Vieta's formula on f(x). The original function f(x) is a cubic: $x^3 - 49x^2 + 623x - 2015$. The sum of its roots is $-(-49)/1 = 49$. Therefore, the sum of the roots of f(x) is 49. Then, the sum of the roots of g(x) would be 49 - 15 = 34. Therefore, the answer is 34?\newline 
    Wait, but let me check this again. If the roots of f(x) are r1, r2, r3, then the roots of
    \\ \midrule
    & \textbf{CoSD with $\alpha=0.5, \beta=0.5 $:} \hlg{Okay}, so \hlg{I need} to find the \hlg{sum} of the roots of $g(x$\hlg{$)$}, \hlg{where} $g(x) = f(x + 5)$. \hlg{The function} $f(x)$ is defined as $f(x) = x^3 - 49x^2 + 623x - 2015$. \newline 
    First, I need to find the function $g(x)$. \hlg{Since} $g(x) = f(x + 5)$, I can substitute $(x + 5)$ \hlg{into} the function $f(x)$ \hlg{in} place of $x$. \hlg{That} gives me $g(x) = (x + 5)^3 - 49(x + 5)^2 + 623(x + 5) - 2015$. \newline 
    \hlg{But maybe there}'s a \hlg{smarter} way to \hlg{find} the sum of the roots of $g(x)$. \hlg{I remember} that \hlg{for} a \hlg{cubic polynomial} of the form $ax^3 + bx^2 + cx + d$, the sum of the roots is $-\frac{b}{a}$. \hlg{So if} I can find the coefficients of the cubic polynomial $g(x)$, I can find the sum of the roots \hlg{without expanding} the polynomial. \newline 
    \hlg{Alternatively}, I can use the fact that \hlg{if} $r$ is a root of $f(x)$, then $r - 5$ is a root of $g(x)$. \hlg{Wait}, \hlg{let} me think about this for a second. If $r$ is a root of $f(x)$, then $f(r) = 0$. \hlg{Then} $f(r - 5) = f(r) = 0$\hlg{?} No, that's not right. \hlg{Wait}, I \hlg{think} I \hlg{need} to \hlg{reverse} the \hlg{substitution}. If $r$ is a root of $f(x)$, then $f(r) = 0$. \hlg{But} $f(r) = g(r - 5)$. \hlg{Because} $g(x) = f(x + 5)$. So if $r$ is a root of $f(x)$, then $g(r - 5) = 0$. \hlg{Therefore}, $r - 5$ is a root of $g(x)$. \hlg{Therefore}, the sum of the roots of $g(x)$ is the sum of the roots of $f(x)$ minus $5$. \newline 
    \hlg{Wait}, \hlg{that}'s not \hlg{quite} right. \hlg{If} $r$ is a root of $f(x)$, then $f(r) = 0$. \hlg{Then} $f(r) = g(r) = 0$. \hlg{No}, that's not right.
    \\ \midrule
    & \textbf{\methodtrain with $\rho=25\%$ and segment size 16:} First, we note that $g(x) = f(x + 5) = (x + 5)^3 - 49(x + 5)^2 + 623(x + 5) - 2015.$  We can expand this expression to \hlg{find the polynomial in standard form, but that would be tedious.  Instead,} we can use the fact that the roots of $g$ are the values of $x$ for which \hlg{$x + 5$ is a root of $f$.  Therefore, if $r_1, r_2, r_3$ are the} roots of $f,$ then $r_1 - 5, r_2 - 5, r_3 - 5$ are the roots of $g.$  By Vieta's formulas, the sum of the roots of $g$ is $(r_1 - 5) + (r_2 - 5) + (r_3 - 5) = (r_1 + r_2 + r_3) - 15.$  Since $r_1 + r_2 + r_3$ is the sum of the roots of $f,$ we can use Vieta's formulas again to find that $r_1 + r_2 + r_3 = 49.$  Therefore, the sum of the roots of $g$ is $49 - 15 = \boxed{34}.$\\[2pt]
    \bottomrule
    \end{tabularx}
    
    \caption{Case study 1 between single models (generator and mentor), CoSD \citep{wang2025speculate} and \methodtrain. We highlight tokens that the mentor inspires the generator in green.}
    \label{tab:case_study_1}
    \end{table*}

    \begin{table*}[t]
        \centering
        \small
        \setlength{\tabcolsep}{4pt}
        \renewcommand{\arraystretch}{0}
        \begin{tabularx}{\linewidth}{@{}>{\centering\arraybackslash}m{0\linewidth} X@{}}
        \toprule
        \multirow{5}{*} &
        \textbf{Question:} Who is the author of 'Under the roofs of Shanghai'?\newline 
        (A) Yang Hansheng (B) Zhou Yang (C) Tian Han (D) Xia Yan
        \\ \midrule
        & \textbf{Generator:} The correct answer is A) Yang Hansheng. \newline 
        Yang Hansheng is a Chinese author, and 'Under the Roofs of Shanghai' is a novel written by him. \newline 
        The novel is a historical fiction that tells the story of the Shanghai International Settlement during the 1920s and 1930s. \newline 
        It is a classic of Chinese literature and has been widely translated and published around the world. \newline
        The novel is known for its vivid portrayal of life in Shanghai during that time period and its exploration of themes such as colonialism, nationalism, and social change.\newline 
        Answer: A. 
        \\ \midrule
        & \textbf{Mentor:} Answer: \newline 
        Okay, let's tackle this question. The user is asking for the author of the book "Under the Roofs of Shanghai." The options are A) Yang Hansheng, B) Zhou Yang, C) Tian Han, D) Xia Yan.\newline 
        First, I need to recall or find out who wrote this book. I know that "Under the Roofs of Shanghai" is a well-known Chinese novel. From what I remember, Xia Yan is a famous Chinese writer, and he's associated with this work. Let me think again. Xia Yan was a prominent figure in Chinese literature, and his works often depicted the lives of people in Shanghai during the early 20th century. "Under the Roofs of Shanghai" (also translated as "Shanghai Roofs") is one of his notable works. \newline 
        Looking at the other options: Yang Hansheng might be a different author, perhaps in a different genre. Zhou Yang and Tian Han are also names I associate with Chinese literature, but I don't recall them being the author of this specific book. Tian Han was a playwright, and Zhou Yang might be a different writer. \newline So, based on this reasoning, the correct answer should be D) Xia Yan.\newline
        Answer: D.
        \\ \midrule
        & \textbf{CoSD with $\alpha=0.5, \beta=0.5 $:} \hlg{Answer}: A.
        \\ \midrule
        & \textbf{\methodtrain with $\rho=25\%$ and segment size 16:} The correct answer is \hlg{D) Xia Yan. 'Under the Roofs of Shanghai' is a novel} by Xia Yan, a Chinese writer. The novel is a historical fiction work that tells the story of the Shanghai International Settlement during the 1920s and 1930s. It is considered a classic of Chinese literature and has been translated into several languages.\newline 
        Answer: D.\\[2pt]
        \bottomrule
        \end{tabularx}
        
        \caption{Case study 2 between single models (generator and mentor), CoSD \citep{wang2025speculate} and \methodtrain. We highlight tokens that the mentor inspires the generator in green.}
        \label{tab:case_study_2}
        \end{table*}
\section{Use of LLMs}
During the preparation of this manuscript, we used Google’s Gemini 3.0 Pro model for minor language refinement and smoothing of the writing. 
The AI tool was not used for generating original content, conducting data analysis, or formulating core scientific ideas. 
All conceptual development, experimentation, and interpretation were conducted independently without reliance on AI tools.

\end{document}